\documentclass[5p,times]{elsarticle}

\usepackage{amssymb}
\usepackage{amsmath}
\usepackage{hyperref}
\usepackage{url}
\usepackage{graphicx}
\usepackage{algorithmic}
\usepackage{textcomp}
\usepackage{xcolor}
\usepackage{booktabs}
\usepackage{makecell}
\usepackage{rotating}
\usepackage{multirow}
\usepackage{subcaption}
\usepackage{wrapfig}

\usepackage{microtype}
\begin{document}

\begin{frontmatter}



\title{CLAIR: Consistent LLM Adaptation via Index-free Reprogramming}

% Equal first authors
\author[aff1]{Dongkuk Kim\fnref{equal}}
\author[aff1]{Jungkyu Shin\fnref{equal}}

% Co-authors
\author[aff2]{Yoonhyuk Choi}
\author[aff3]{Jong Wook Kim}

% Corresponding author
\author[aff1]{Beakcheol Jang\corref{cor1}}

% Author footnotes
\fntext[equal]{These authors contributed equally to this work.}
\cortext[cor1]{Corresponding author.}

% Affiliations
\affiliation[aff1]{
    organization={Yonsei University},
    city={Seoul},
    country={Republic of Korea}
}

\affiliation[aff2]{
    organization={Sookmyung Women's University},
    city={Seoul},
    country={Republic of Korea}
}

\affiliation[aff3]{
    organization={Sangmyung University},
    city={Seoul},
    country={Republic of Korea}
}

% %% use optional labels to link authors explicitly to addresses:
% %% \author[label1,label2]{}
% %% \affiliation[label1]{organization={},
% %%             addressline={},
% %%             city={},
% %%             postcode={},
% %%             state={},
% %%             country={}}
% %%
% %% \affiliation[label2]{organization={},
% %%             addressline={},
% %%             city={},
% %%             postcode={},
% %%             state={},
% %%             country={}}

% \author{} %% Author name

% \affiliation{organization={},%Department and Organization

%% Abstract
\begin{abstract}
Time-series forecasting with pretrained large language models
(LLMs) uses contextualized patch representations, raising the
question of how to align them.
Matching patches at the same temporal position is natural, but
self-attention allows each patch state to incorporate information
from elsewhere in the input window, suggesting that useful
correspondences can extend across positions.
This flexibility, however, does not necessarily translate into
consistent forecasts: similar patch states may play different
roles when used in their original temporal order.
To address both issues, we propose CLAIR (Consistent LLM
Adaptation via Index-free Reprogramming), a teacher-guided
framework that combines patch-level transport alignment (PTA)
and forecast-level consistency (FC).
A clean-view exponential moving average (EMA) teacher guides
a mildly perturbed student, with PTA establishing soft,
representation-dependent correspondences and FC encouraging
agreement at corresponding future steps.
The forecasting head retains the original patch order.
We test this combination through controlled comparisons:
while PTA and index-wise alignment perform similarly without
FC, PTA+FC under EMA supervision reduces horizon-averaged
mean squared error (MSE) by 2.1\% on ETTh1 and 2.6\% on
ETTm1 relative to Index+FC.
The same preference for PTA+FC emerges in comparisons with
alternative matching strategies on Weather and ETTh2.
Beyond these controls, CLAIR achieves the lowest
horizon-averaged MSE on five of six long-term benchmarks
and the lowest aggregate overall weighted average (OWA)
on M4 among evaluated methods, with competitive few-shot
performance.
As the teacher and transport are used only during training,
inference requires only the student.
\end{abstract}

%% Keywords

\end{frontmatter}



\section{Introduction}

Time-series forecasting supports decision-making in domains such as energy, traffic, finance, climate, healthcare, and industrial monitoring.
In these settings, time-series data contain noise, exhibit non-stationary dynamics, and involve dependencies spanning multiple temporal scales.
These characteristics make it challenging to learn representations that remain informative for future predictions.
Recent work has explored adapting pretrained large language models (LLMs) as forecasting backbones to leverage their capacity for modeling contextual dependencies in time-series forecasting through patch-based input representations~\citep{Zhou2023OneFitsAll,Jin2024TIMELLM}.
These efforts have further expanded toward learning and leveraging backbone-derived patch representations through feature- and output-level supervision and temporal, semantic, or structural conditioning~\citep{Pan2024S2IP,Hu2025FSCA,Liu2026SELLM,Liu2025CALF,Wang2025TSHTFA}.
In particular, when contextualized patch representations
are used for alignment-based supervision, an important
design consideration is how temporal position should
guide correspondence between patch states, given that
each state may incorporate information from other
temporal positions.

\begin{figure*}[t]
    \centering
    \includegraphics[width=\textwidth]
    {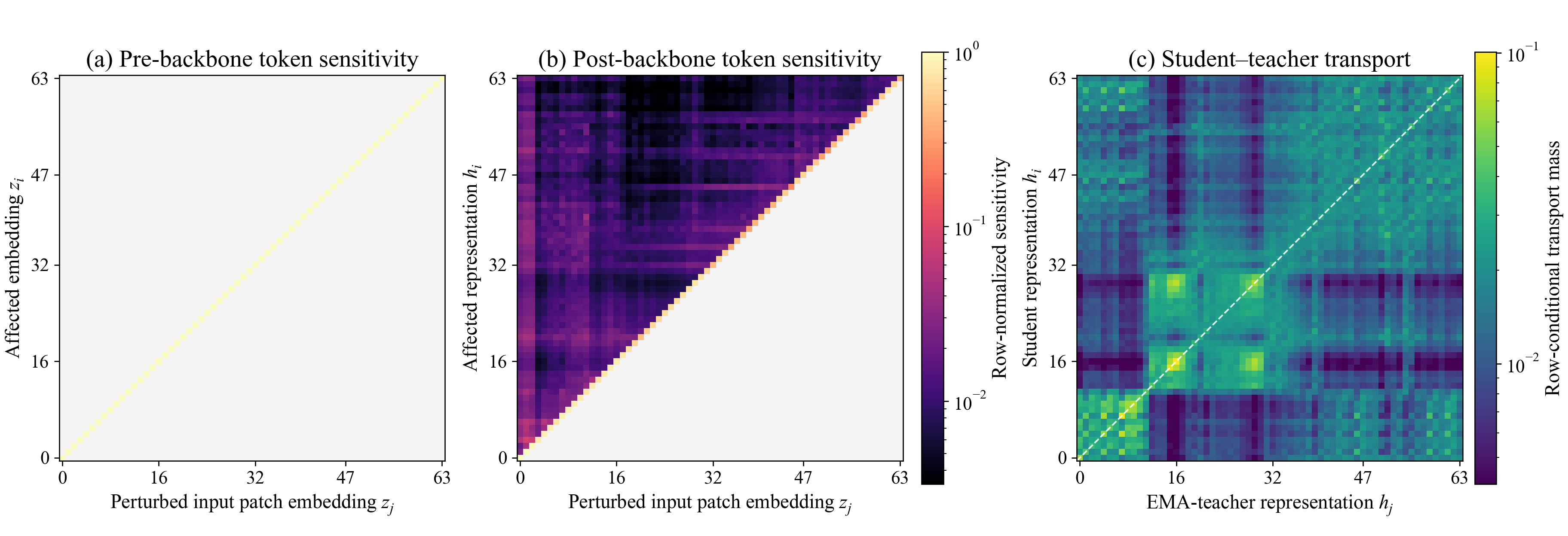}
    \caption{Contextualization of patch representations.
(a, b) Sensitivity of patch representation at position $i$ (y-axis) to replacing the input patch at position $j$ (x-axis).
(a) Changes are localized before the backbone, yielding a diagonal pattern.
(b) After self-attention, changes extend across temporal positions, reflecting contextualization beyond individual segments.
(c) Learned student--teacher patch correspondence in CLAIR for the same sequence, including associations beyond same-index matching (dashed diagonal).
}
    \label{fig:token_sensitivity_transport}
\end{figure*}

The first consideration concerns whether correspondence between contextualized patch representations should be restricted to the same temporal positions.
When patch representations share the same temporal grid, index-wise alignment remains a natural reference because patches at the same position correspond to the same observed interval.
However, whether this positional correspondence captures all potentially useful relationships between contextualized patch representations needs further examination.
Similar temporal patterns can recur at different positions within an input window, and useful relationships may therefore exist between patch representations at different positions.
This possibility is particularly relevant when aligning post-backbone representations, as each patch state can incorporate contextual information from other temporal positions through self-attention~\citep{Vaswani2017Attention}.
Indeed, our token-replacement diagnostic illustrates this contextualization: changes localized to a single position before the backbone extend to representations at other positions afterward (Figure~\ref{fig:token_sensitivity_transport}(a,b)).
These observations motivate examining whether soft, representation-dependent correspondence across positions can provide additional supervision beyond same-index matching.

A second consideration concerns how such flexible representation alignment relates to forecasting.
Even when patch states at different temporal positions are similar or aligned, they need not play identical roles in predicting future values.
Because the forecasting head processes patch states in their original temporal order, aligning representations across positions does not alter their temporal roles or necessarily ensure agreement at corresponding future time steps.

To address these complementary considerations, we propose CLAIR, an asymmetric teacher-guided framework for Consistent LLM Adaptation via Index-free Reprogramming.
CLAIR employs a clean-view EMA teacher and a perturbed-view student, using patch-level transport alignment (PTA) to establish soft, representation-dependent correspondences between their contextualized patch representations through optimal transport, rather than restricting alignment to same-index pairs.
Forecast-level consistency (FC) complements this flexible alignment by encouraging agreement between teacher and student predictions at corresponding future time steps, while preserving the original temporal order for forecasting.
Our main contributions are summarized as follows:
\begin{itemize}
    \item We introduce CLAIR, an asymmetric teacher--student
    framework that couples flexible PTA with FC under a clean-view EMA teacher, while preserving the
    original temporal order for forecasting. The teacher and transport are used only during training,
    enabling student-only inference.

    \item Controlled comparisons show that PTA and index-wise
    alignment perform comparably without FC, whereas PTA
    achieves lower forecasting errors when combined with FC,
    particularly under EMA supervision. Comparisons with
    alternative correspondence strategies further support
    the effectiveness of flexible patch alignment.

    \item Evaluations across long-term, short-term, and few-shot
    forecasting demonstrate competitive performance against
    representative baselines, including the lowest
    horizon-averaged MSE on five of six long-term benchmarks
    and the lowest aggregate OWA on M4.
    
\end{itemize}

\section{Related Work}

\subsection{Teacher--Student and Non-contrastive Learning}

Teacher--student learning encompasses both prediction consistency and representation-level supervision. Weight-averaged teacher predictions provide consistency targets under perturbations~\citep{Tarvainen2017MeanTeacher}. At the representation level, non-contrastive approaches learn across views without explicit negative pairs, using asymmetric branches with momentum targets or stop-gradient and a prediction head~\citep{Grill2020BYOL,Chen2021SimSiam}. These approaches differ in both how reference targets are updated and whether the learning signal acts on representations or final predictions. In time-series learning, teacher supervision has been incorporated into hierarchical masked modeling for forecasting~\citep{Zhao2024himtm}, reconstruction-based representation learning~\citep{Skaf2025tsrc}, and change-point modeling~\citep{Fan2025stas}. Beyond target construction, distillation also requires specifying the correspondence between the quantities being matched. Language-model distillation illustrates different choices: predefined layer and token-position mappings support hidden-state matching~\citep{Jiao2020TinyBERT}, whereas optimal transport enables token- and sequence-level alignment of logit distributions across different tokenizers~\citep{Cui2025MultiLevelOT}.

\begin{figure*}[t!]
    \centering
    \includegraphics[width=\textwidth]{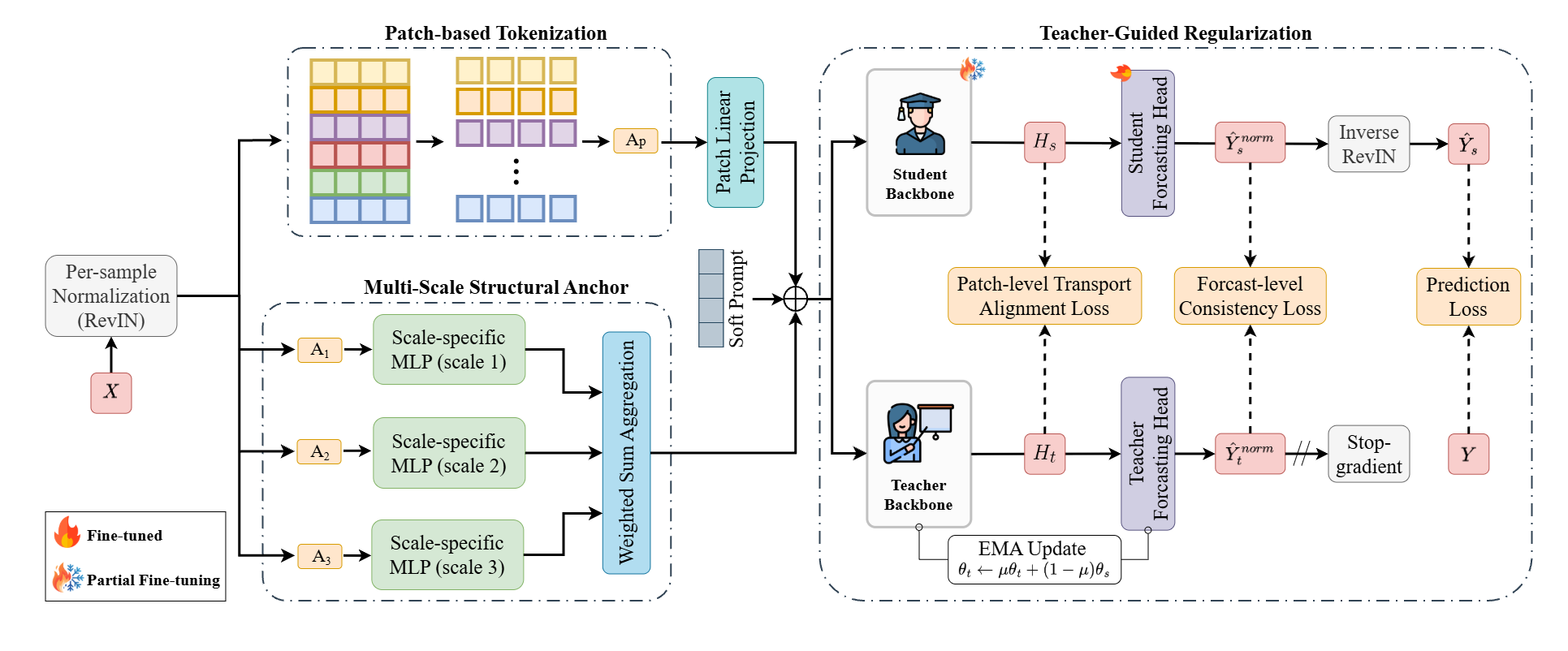}
    \caption{
    Overall architecture of CLAIR.
    Patch tokens are combined with soft prompts and a multi-scale
    structural anchor.
    $\mathcal{A}_p$ and $\mathcal{A}_k$ ($k=1,2,3$) apply Gaussian
    noise and elementwise dropout to extracted patch and scale-$k$
    anchor windows, respectively.
    Perturbations are sampled independently for patch windows and
    each anchor scale and applied only to the student during training.
    The student and EMA teacher use separate copies of the input encoders.
    The teacher guides the student through patch-level transport
    alignment and forecast-level consistency.
    Only the student is used at inference.
    }
    \label{fig:clair_arch}
\end{figure*}

\subsection{LLM-based Time-Series Forecasting and Adaptation}
LLM-based time-series forecasting adapts pretrained language models through numerical input interfaces and task-specific parameter updates. OFA projects time-series patches into the embedding space and selectively updates pretrained parameters~\citep{Zhou2023OneFitsAll}, while LLM4TS also employs patch-based adaptation~\citep{Chang2025LLM4TS}. Time-LLM reprograms patches through text prototypes and incorporates natural-language prefix prompts~\citep{Jin2024TIMELLM}. S$^2$IP-LLM derives semantic anchors from pretrained word embeddings for alignment and prompting~\citep{Pan2024S2IP}. Structural and temporal modeling further enrich these interfaces. FSCA uses graph-based context alignment to model structural and logical relations between time-series tokens and language prompts, with modules that can be integrated into multiple LLM layers~\citep{Hu2025FSCA}. SE-LLM enhances temporal--semantic representations and introduces an attention-integrated Time-Adapter for temporal dependency modeling~\citep{Liu2026SELLM}. Cross-modal approaches also differ in the representations and objectives used for supervision. CALF regularizes intermediate features and outputs across temporal and language-space branches~\citep{Liu2025CALF}. TimeCMA aligns time-series representations with LLM-derived prompt embeddings~\citep{Liu2025TimeCMA}, while TS-HTFA combines intermediate-feature alignment with output-level optimal transport~\citep{Wang2025TSHTFA}. Whereas these alignment-oriented approaches primarily differ in the representations and supervision objectives, CLAIR treats patch correspondence itself as an explicit design choice. Specifically, it establishes flexible patch-level transport between clean and perturbed views of the same series, while examining how flexible versus index-wise alignment interacts with FC.

\section{Method}

CLAIR couples PTA with FC under a shared clean-view EMA teacher, guiding a student whose extracted patch and anchor windows are mildly perturbed.
Soft prompts and a multi-scale structural anchor provide task-level and sequence-specific input conditioning, respectively.
Figure~\ref{fig:clair_arch} presents an overview of the architecture.

\subsection{Problem Setting}

Given multivariate input sequences $X$, the goal is to generate forecasts $\hat{Y}$ for the ground-truth future sequences $Y$, where
\begin{equation}
    X \in \mathbb{R}^{B \times L \times M},
    \qquad
    Y,\hat{Y} \in \mathbb{R}^{B \times H \times M}.
\end{equation}
Here, $B$, $L$, $M$, and $H$ denote the batch size, input length,
number of variables, and forecasting horizon, respectively.
Following the channel-independent setting, we reshape $X$ into
$U=BM$ univariate sequences $x^{(u)} \in \mathbb{R}^{L}$ and
normalize each using its input-window statistics.
Predictions are inverse-normalized using the same statistics
and reshaped into the multivariate output format.

\subsection{CLAIR Architecture}

\paragraph{Backbone and Patch Reprogramming}
We use GPT-2~\citep{radford2019language} as the pretrained backbone.
Its self-attention and feed-forward parameters are frozen, while its positional embeddings and layer-normalization parameters are updated together with the patch projection, input-conditioning modules, and forecasting head.
Each normalized sequence $\bar{x}^{(u)}$ is divided into $N$ overlapping patches of length $P$ and stride $S$.
For each patch $p_i^{(u)}\in\mathbb{R}^{P}$, a shared linear layer projects it into the GPT-2 embedding space:
\begin{equation}
    z_i^{(u)}
    =
    W_p p_i^{(u)} + b_p,
    \qquad
    W_p\in\mathbb{R}^{D\times P},
    \quad
    b_p,z_i^{(u)}\in\mathbb{R}^{D},
    \label{eq:patch_projection}
\end{equation}
where $D$ is the GPT-2 hidden dimension.
The projected patch tokens are stacked as
\begin{equation}
    Z_{\mathrm{patch}}^{(u)}
    =
    [z_1^{(u)},\ldots,z_N^{(u)}]^\top
    \in
    \mathbb{R}^{N\times D}.
\end{equation}

\paragraph{Input Conditioning}
Learnable soft prompts
$P_{\mathrm{soft}} \in \mathbb{R}^{n_p \times D}$
provide an input-independent task prefix shared across sequences.
In contrast, the structural anchor
$a_{\mathrm{struct}}^{(u)} \in \mathbb{R}^{D}$
summarizes the current input across $K$ temporal scales into
a single context token.
At scale $k$, a two-layer MLP $f_k(\cdot)$ encodes sliding
windows $w_i^{(u,k)}$, whose features are aggregated as
\begin{equation}
\begin{aligned}
    s^{(u,k)}
    &=
    \frac{\alpha_k}{T_k}
    \sum_{i=1}^{T_k}
    f_k\!\left(w_i^{(u,k)}\right), \\
    a_{\mathrm{struct}}^{(u)}
    &=
    \sum_{k=1}^{K}\pi_k s^{(u,k)},
    \qquad
    \pi_k =
    \frac{\exp(\beta_k)}
    {\sum_{j=1}^{K}\exp(\beta_j)}.
\end{aligned}
\label{eq:structural_anchor}
\end{equation}
Here, $T_k$ is the window count, $\alpha_k$ rescales each
scale-level summary, and $\pi_k$ provides normalized mixing
across scales.
The gains $\alpha_k$ and logits $\beta_k$ are learned and
shared across sequences.

The prompts and anchor are prepended to the patch embeddings:
\begin{equation}
Z_{\mathrm{in}}^{(u)}
=
\left[
    P_{\mathrm{soft}}
    \parallel
    a_{\mathrm{struct}}^{(u)}
    \parallel
    Z_{\mathrm{patch}}^{(u)}
\right]
\in
\mathbb{R}^{(n_p+1+N)\times D},
    \label{eq:gpt_input}
\end{equation}
where $\parallel$ denotes concatenation along the token dimension.
These prefixes condition patch representations through GPT-2
attention.
Only the patch hidden states
$H_{\mathrm{patch}} \in \mathbb{R}^{U \times N \times D}$
are retained for PTA and the forecasting head $g$, which produces
normalized forecasts:
\begin{equation}
    \hat{Y}^{\mathrm{norm}}
    =
    g\!\left(
        \operatorname{flatten}(H_{\mathrm{patch}})
    \right),
    \qquad
    \hat{Y}^{\mathrm{norm}} \in \mathbb{R}^{U \times H}.
    \label{eq:forecasting_head}
\end{equation}
Flattening combines the patch and feature dimensions within
each sequence in their original order, which is unchanged by PTA. 

\subsection{Teacher-guided Regularization}

\paragraph{Asymmetric EMA Teacher}
The student $F_{\theta_s}$ and teacher $F_{\theta_t}$ share the same architecture.
The teacher is initialized from the student and updated after each student optimization step as
\begin{equation}
    \theta_t
    \leftarrow
    \mu\theta_t
    +
    (1-\mu)\theta_s,
    \label{eq:ema_update}
\end{equation}
where $\mu$ is the EMA momentum.
The update covers all trainable student parameters, including the forecasting head and the unfrozen GPT-2 parameters.
The teacher receives no gradients, and its outputs are treated as stop-gradient targets.

For each normalized sequence, let
$P^{(u)}=[p_1^{(u)},\ldots,p_N^{(u)}]$
denote the patch windows and
$W^{(u,k)}=[w_1^{(u,k)},\ldots,w_{T_k}^{(u,k)}]$
the windows at anchor scale $k$.
The two branches process asymmetric versions of these extracted windows:
\begin{equation}
\begin{aligned}
    P_s^{(u)}
    &=
    \mathcal{A}_p\!\left(P^{(u)}\right),
    &
    P_t^{(u)}
    &=
    P^{(u)}, \\
    W_s^{(u,k)}
    &=
    \mathcal{A}_k\!\left(W^{(u,k)}\right),
    &
    W_t^{(u,k)}
    &=
    W^{(u,k)},
    \qquad k=1,\ldots,K .
\end{aligned}
\label{eq:asymmetric_views}
\end{equation}
Here, $\mathcal{A}_p$ and $\mathcal{A}_k$ denote stochastic perturbation operators applied independently to the extracted patch windows and the windows at anchor scale $k$, respectively. Specifically, these operators apply Gaussian noise and elementwise dropout only to the student branch, before patch projection and anchor encoding.
The corresponding perturbed and unperturbed windows are then processed by the student and teacher, producing
$H_s,\hat{Y}_s^{\mathrm{norm}}$ and
$H_t,\hat{Y}_t^{\mathrm{norm}}$, respectively.
Here,
$H_s,H_t \in \mathbb{R}^{U \times N \times D}$
and
$\hat{Y}_s^{\mathrm{norm}},
\hat{Y}_t^{\mathrm{norm}}
\in \mathbb{R}^{U \times H}$.
The clean view refers to the unperturbed extracted windows, rather than to an assumption that the original observations are noise-free.

\paragraph{Patch-level Transport Alignment}
Although the two branches retain the same patch order, their contextualized post-backbone representations need not be compared only through fixed same-index correspondence.
PTA therefore permits flexible representation-level correspondence.

We first normalize the patch representations along the embedding dimension:
\begin{equation}
    \tilde{H}_s
    =
    \operatorname{normalize}(H_s),
    \qquad
    \tilde{H}_t
    =
    \operatorname{normalize}(H_t).
    \label{eq:representation_normalization}
\end{equation}
For each sequence $u$, the pairwise cost matrix is
\begin{equation}
    C_{ij}^{(u)}
    =
    \left\|
        \tilde{h}_{s,i}^{(u)}
        -
        \operatorname{sg}
        \left(
            \tilde{h}_{t,j}^{(u)}
        \right)
    \right\|_2^2,
    \label{eq:pta_cost}
\end{equation}
where $\operatorname{sg}(\cdot)$ denotes stop-gradient.

With uniform marginals
$r=c=\frac{1}{N}\mathbf{1}_N$,
the transport plan is obtained by
\begin{equation}
    \Pi^{\star(u)}
    =
    \arg\min_{\Pi \in \mathcal{U}(r,c)}
    \left\langle
        \Pi,
        C^{(u)}
    \right\rangle
    +
    \varepsilon_{\mathrm{OT}}
    \sum_{i,j}
    \Pi_{ij}
    \left(
        \log\Pi_{ij}-1
    \right),
    \label{eq:pta_plan}
\end{equation}
where $\mathcal{U}(r,c)$ is the set of transport plans with marginals $r$ and $c$, and
$\varepsilon_{\mathrm{OT}}$ is the entropic regularization coefficient.
The transport plan is computed using Sinkhorn iterations~\citep{Cuturi2013Sinkhorn}.

The PTA loss is
\begin{equation}
    \mathcal{L}_{\mathrm{PTA}}
    =
    \frac{1}{U}
    \sum_{u=1}^{U}
    \left\langle
        \Pi^{\star(u)},
        C^{(u)}
    \right\rangle .
    \label{eq:pta_loss}
\end{equation}
Importantly, the transport plan is used only to define correspondence in the representation-level loss. It does not reorder or aggregate the hidden states passed to the forecasting head.

\paragraph{Forecast-level Consistency}
Because PTA permits flexible correspondence at the representation level, a low transport cost alone does not require the forecasts produced from the ordered hidden states to agree.
FC therefore provides a complementary forecast-level constraint on the student and teacher predictions.
Applying the corresponding forecasting heads in \eqref{eq:forecasting_head} to $H_s$ and $H_t$ produces
$\hat{Y}_s^{\mathrm{norm}}$ and
$\hat{Y}_t^{\mathrm{norm}}$, respectively:
\begin{equation}
    \mathcal{L}_{\mathrm{FC}}
    =
    \frac{1}{UH}
    \left\|
        \hat{Y}_s^{\mathrm{norm}}
        -
        \operatorname{sg}
        \left(
            \hat{Y}_t^{\mathrm{norm}}
        \right)
    \right\|_F^2 .
    \label{eq:fc_loss}
\end{equation}

\subsection{Optimization and Inference}

The inverse-normalized student forecast
$\hat{Y}_s \in \mathbb{R}^{B \times H \times M}$
is optimized using a task-specific supervised forecasting loss:
\begin{equation}
    \mathcal{L}_{\mathrm{pred}}
    =
    \ell_{\mathrm{pred}}
    \left(
        \hat{Y}_s,
        Y
    \right),
    \label{eq:prediction_loss}
\end{equation}
where $\ell_{\mathrm{pred}}$ denotes the supervised forecasting criterion used for the corresponding forecasting setting.

The final objective is
\begin{equation}
    \mathcal{L}
    =
    \mathcal{L}_{\mathrm{pred}}
    +
    \gamma_e
    \left(
        \lambda_{\mathrm{PTA}}
        \mathcal{L}_{\mathrm{PTA}}
        +
        \lambda_{\mathrm{FC}}
        \mathcal{L}_{\mathrm{FC}}
    \right),
    \label{eq:training_objective}
\end{equation}
where $\lambda_{\mathrm{PTA}}$ and $\lambda_{\mathrm{FC}}$ control the two auxiliary objectives, and
\begin{equation}
    \gamma_e
    =
    \min\!\left(
        1,
        \frac{e+1}{E_{\mathrm{warm}}}
    \right)
    \label{eq:loss_warmup}
\end{equation}
gradually introduces them during the first
$E_{\mathrm{warm}}$ epochs.
At inference, only the student model is retained.
Student-side perturbation, the EMA teacher, Sinkhorn transport, and the auxiliary consistency losses are removed.

\begin{table*}[!t]
\centering
\caption{
Long-term forecasting results averaged over prediction horizons $H$.
}
\label{tab:long-term-forcasting_results}

\resizebox{\textwidth}{!}{%
\begin{tabular}{l*{10}{cc}}
\toprule
\multicolumn{1}{c}{}
& \multicolumn{2}{c}{CLAIR}
& \multicolumn{2}{c}{Time-LLM}
& \multicolumn{2}{c}{S\textsuperscript{2}IP-LLM}
& \multicolumn{2}{c}{OFA}
& \multicolumn{2}{c}{CALF}
& \multicolumn{2}{c}{FSCA}
& \multicolumn{2}{c}{SE-LLM}
& \multicolumn{2}{c}{DLinear}
& \multicolumn{2}{c}{iTransformer}
& \multicolumn{2}{c}{PatchTST} \\
\cmidrule(lr){2-3}
\cmidrule(lr){4-5}
\cmidrule(lr){6-7}
\cmidrule(lr){8-9}
\cmidrule(lr){10-11}
\cmidrule(lr){12-13}
\cmidrule(lr){14-15}
\cmidrule(lr){16-17}
\cmidrule(lr){18-19}
\cmidrule(lr){20-21}
Dataset
& MSE & MAE
& MSE & MAE
& MSE & MAE
& MSE & MAE
& MSE & MAE
& MSE & MAE
& MSE & MAE
& MSE & MAE
& MSE & MAE
& MSE & MAE \\
\midrule

Weather
& \textbf{0.2261} & \textbf{0.2642}
& 0.2346 & 0.2724
& 0.2284 & \underline{0.2658}
& 0.2285 & 0.2667
& 0.2363 & 0.2719
& \underline{0.2281} & 0.2678
& 0.2314 & 0.2676
& 0.2416 & 0.2941
& 0.2443 & 0.2775
& 0.2338 & 0.2749 \\

ETTh1
& \textbf{0.4094} & \textbf{0.4287}
& 0.4409 & 0.4517
& \underline{0.4154} & 0.4346
& 0.4265 & 0.4356
& 0.4281 & 0.4434
& 0.4180 & \underline{0.4332}
& 0.4258 & 0.4379
& 0.4184 & 0.4386
& 0.4485 & 0.4612
& 0.4600 & 0.4643 \\

ETTh2
& \textbf{0.3564} & \textbf{0.3971}
& 0.3686 & 0.4056
& 0.3624 & 0.4052
& 0.3742 & 0.4065
& 0.3666 & 0.4020
& \underline{0.3597} & \underline{0.3993}
& 0.3823 & 0.4113
& 0.4995 & 0.4803
& 0.3846 & 0.4154
& 0.4036 & 0.4305 \\

ETTm1
& \textbf{0.3493} & \underline{0.3820}
& 0.3790 & 0.3966
& \underline{0.3509} & 0.3835
& 0.3596 & 0.3899
& 0.3580 & 0.3870
& 0.3674 & 0.3948
& 0.3684 & 0.3948
& 0.3586 & \textbf{0.3816}
& 0.3689 & 0.3989
& 0.3770 & 0.4048 \\

ETTm2
& \textbf{0.2563} & \textbf{0.3200}
& 0.2649 & 0.3273
& 0.2599 & 0.3236
& 0.2663 & 0.3288
& 0.2628 & \underline{0.3203}
& \underline{0.2581} & 0.3212
& 0.2765 & 0.3280
& 0.2768 & 0.3422
& 0.2718 & 0.3312
& 0.2797 & 0.3353 \\

Electricity
& \underline{0.1639} & \underline{0.2584}
& 0.1719 & 0.2720
& 0.1687 & 0.2660
& 0.1664 & 0.2630
& 0.1664 & 0.2611
& \textbf{0.1635} & \textbf{0.2574}
& 0.1688 & 0.2627
& 0.1671 & 0.2675
& 0.1640 & \underline{0.2584}
& 0.1663 & 0.2664 \\

\bottomrule
\end{tabular}%
}
\end{table*}
\begin{table*}
\centering
\caption{Short-term forecasting results on the M4 dataset.}
\label{tab:short_term_forcasting_results}

\resizebox{\textwidth}{!}{
    % \begin{tabular}{c|l|cccccccccc}
    \begin{tabular}{clcccccccccc}
    \toprule
    % \multicolumn{2}{c|}{Models}
        \multicolumn{2}{c}{}
    & CLAIR
    & Time-LLM
    & S\textsuperscript{2}IP-LLM
    & OFA
    & CALF
    & FSCA
    & SE-LLM
    & DLinear
    & iTransformer
    & PatchTST
     \\
    \midrule

    \multirow{3}{*}{Average}
    & SMAPE
    & \underline{12.008}
    & 12.522
    & 12.314
    & 12.256
    & 12.193
    & \textbf{12.003}
    & 12.145
    & 12.507
    & 13.359
    & 12.873
     \\

    & MASE
    & \textbf{1.628}
    & 1.704
    & 1.684
    & 1.714
    & \underline{1.632}
    & 1.637
    & 1.656
    & 1.684
    & 1.786
    & 1.757
     \\

    & OWA
    & \textbf{0.868}
    & 0.907
    & 0.894
    & 0.900
    & 0.876
    & \underline{0.871}
    & 0.881
    & 0.901
    & 0.959
    & 0.934
     \\

    \bottomrule
    \end{tabular}
}
\end{table*}
%==================== Few-shot 10% ====================%
\begin{table*}[!t]
\centering
\caption{
Few-shot forecasting results with 10\% training data averaged over prediction horizons $H$.
}
\label{tab:fewshot10_results}

\resizebox{\textwidth}{!}{
    % \begin{tabular}{l|cc|cc|cc|cc|cc|cc|cc|cc|cc|cc}
    \begin{tabular}{lcccccccccccccccccccc}
    \toprule
    \multicolumn{1}{c}{}
    & \multicolumn{2}{c}{CLAIR}
    & \multicolumn{2}{c}{Time-LLM}
    & \multicolumn{2}{c}{S\textsuperscript{2}IP-LLM}
    & \multicolumn{2}{c}{OFA}
    & \multicolumn{2}{c}{CALF}
    & \multicolumn{2}{c}{FSCA}
    & \multicolumn{2}{c}{SE-LLM}
    & \multicolumn{2}{c}{DLinear}
    & \multicolumn{2}{c}{iTransformer}
    & \multicolumn{2}{c}{PatchTST} \\
    \cmidrule(lr){2-3}
    \cmidrule(lr){4-5}
    \cmidrule(lr){6-7}
    \cmidrule(lr){8-9}
    \cmidrule(lr){10-11}
    \cmidrule(lr){12-13}
    \cmidrule(lr){14-15}
    \cmidrule(lr){16-17}
    \cmidrule(lr){18-19}
    \cmidrule(lr){20-21}

    % \multicolumn{1}{c|}{Metrics}
    \multicolumn{1}{l}{Dataset}
    & MSE & MAE
    & MSE & MAE
    & MSE & MAE
    & MSE & MAE
    & MSE & MAE
    & MSE & MAE
    & MSE & MAE
    & MSE & MAE
    & MSE & MAE
    & MSE & MAE \\
    \midrule

    Weather
    & \textbf{0.2372} & \textbf{0.2766}
    & 0.2491 & 0.2874
    & \underline{0.2404} & \underline{0.2787}
    & 0.2430 & 0.2793
    & 0.2573 & 0.2937
    & 0.2425 & 0.2815
    & 0.2512 & 0.2844
    & 0.2420 & 0.2894
    & 0.2893 & 0.3164
    & 0.2529 & 0.2893 \\

    ETTh1
    & \underline{0.5589} & \underline{0.5064}
    & 0.6879 & 0.5585
    & 0.7432 & 0.5759
    & 0.5912 & 0.5263
    & 0.7168 & 0.5860
    & 0.6337 & 0.5389
    & 0.6420 & 0.5427
    & \textbf{0.5309} & \textbf{0.5025}
    & 0.7027 & 0.5787
    & 0.7259 & 0.5966 \\

    ETTh2
    & \textbf{0.3846} & \textbf{0.4226}
    & 0.4042 & 0.4333
    & 0.4647 & 0.4660
    & 0.4449 & 0.4578
    & 0.4207 & 0.4422
    & 0.4112 & 0.4345
    & \underline{0.3951} & \underline{0.4272}
    & 0.4137 & 0.4459
    & 0.4262 & 0.4473
    & 0.4556 & 0.4659 \\

    ETTm1
    & \underline{0.4482} & \underline{0.4335}
    & 0.4743 & 0.4534
    & 0.4834 & 0.4600
    & 0.4599 & 0.4391
    & 0.4985 & 0.4721
    & 0.4924 & 0.4540
    & 0.6259 & 0.5152
    & \textbf{0.4131} & \textbf{0.4275}
    & 0.5705 & 0.5101
    & 0.6023 & 0.5033 \\

    ETTm2
    & \textbf{0.2611} & \textbf{0.3217}
    & 0.3017 & 0.3515
    & 0.3077 & 0.3531
    & 0.2899 & 0.3358
    & 0.2944 & 0.3427
    & \underline{0.2862} & \underline{0.3341}
    & 0.2892 & 0.3419
    & 0.2965 & 0.3507
    & 0.3147 & 0.3601
    & 0.3299 & 0.3646 \\

    Electricity
    & 0.1819 & 0.2778
    & 0.1860 & 0.2846
    & \underline{0.1788} & \underline{0.2743}
    & \textbf{0.1782} & \textbf{0.2723}
    & 0.1932 & 0.2909
    & 0.1820 & 0.2762
    & 0.1892 & 0.2829
    & 0.2201 & 0.3288
    & 0.1946 & 0.2913
    & 0.1848 & 0.2841 \\

    \bottomrule
    \end{tabular}
}
\end{table*}
%==================== Few-shot 5% ====================%
\begin{table*}[!t]
\centering
\caption{
Few-shot forecasting results with 5\% training data averaged over prediction horizons $H$.
}
\label{tab:fewshot5_results}

\resizebox{\textwidth}{!}{
    % \begin{tabular}{l|cc|cc|cc|cc|cc|cc|cc|cc|cc|cc}
    \begin{tabular}{lcccccccccccccccccccc}
    \toprule
    \multicolumn{1}{c}{}
    & \multicolumn{2}{c}{CLAIR}
    & \multicolumn{2}{c}{Time-LLM}
    & \multicolumn{2}{c}{S\textsuperscript{2}IP-LLM}
    & \multicolumn{2}{c}{OFA}
    & \multicolumn{2}{c}{CALF}
    & \multicolumn{2}{c}{FSCA}
    & \multicolumn{2}{c}{SE-LLM}
    & \multicolumn{2}{c}{DLinear}
    & \multicolumn{2}{c}{iTransformer}
    & \multicolumn{2}{c}{PatchTST} \\
    \cmidrule(lr){2-3}
    \cmidrule(lr){4-5}
    \cmidrule(lr){6-7}
    \cmidrule(lr){8-9}
    \cmidrule(lr){10-11}
    \cmidrule(lr){12-13}
    \cmidrule(lr){14-15}
    \cmidrule(lr){16-17}
    \cmidrule(lr){18-19}
    \cmidrule(lr){20-21}

    % \multicolumn{1}{c|}{Metrics}
    \multicolumn{1}{l}{Dataset}
    & MSE & MAE
    & MSE & MAE
    & MSE & MAE
    & MSE & MAE
    & MSE & MAE
    & MSE & MAE
    & MSE & MAE
    & MSE & MAE
    & MSE & MAE
    & MSE & MAE \\
    \midrule

    Weather
    & \textbf{0.2642} & \textbf{0.3046}
    & 0.2778 & 0.3097
    & 0.2918 & 0.3244
    & 0.2720 & 0.3089
    & 0.2759 & 0.3113
    & \underline{0.2670} & \underline{0.3055}
    & 0.2745 & 0.3087
    & 0.2724 & 0.3245
    & 0.2918 & 0.3273
    & 0.2846 & 0.3180 \\

    ETTh1
    & \underline{0.6719} & \underline{0.5509}
    & 0.8125 & 0.5919
    & 1.0053 & 0.6719
    & 0.8402 & 0.6190
    & 0.8672 & 0.6376
    & 0.7419 & 0.5887
    & 0.7752 & 0.6030
    & \textbf{0.6269} & \textbf{0.5478}
    & 0.8137 & 0.6157
    & 0.8856 & 0.6572 \\

    ETTh2
    & \textbf{0.3729} & \textbf{0.4235}
    & 0.5466 & 0.5100
    & 0.6167 & 0.5454
    & 0.4697 & 0.4844
    & 0.4411 & 0.4591
    & \underline{0.3754} & \underline{0.4248}
    & 0.4600 & 0.4807
    & 0.4872 & 0.5090
    & 0.4313 & 0.4516
    & 0.4052 & 0.4274 \\

    ETTm1
    & \underline{0.4604} & \textbf{0.4452}
    & 0.5077 & 0.4729
    & 0.5201 & 0.4860
    & 0.4939 & 0.4625
    & 0.5242 & 0.4850
    & 0.4970 & 0.4630
    & 0.6061 & 0.5066
    & \textbf{0.4464} & \underline{0.4463}
    & 0.5634 & 0.5047
    & 0.5703 & 0.5047 \\

    ETTm2
    & \textbf{0.2933} & \textbf{0.3405}
    & 0.3228 & 0.3618
    & 0.3199 & 0.3624
    & 0.3174 & 0.3517
    & 0.3197 & 0.3598
    & \underline{0.3064} & \underline{0.3473}
    & 0.3083 & 0.3514
    & 0.3263 & 0.3762
    & 0.3085 & 0.3573
    & 0.3533 & 0.3787 \\

    Electricity
    & \textbf{0.1806} & \textbf{0.2757}
    & 0.1890 & 0.2888
    & 0.1831 & 0.2801
    & 0.1851 & 0.2829
    & 0.2281 & 0.3231
    & \underline{0.1821} & \underline{0.2787}
    & 0.1967 & 0.2911
    & 0.2958 & 0.3935
    & 0.2102 & 0.3090
    & 0.1903 & 0.2901 \\

    \bottomrule
    \end{tabular}
}
\end{table*}

\section{Experiments}
\label{sec:experiments}
We evaluate CLAIR across long-term, short-term, and few-shot forecasting settings.
Controlled comparisons examine PTA and index-wise alignment with and without FC.
Component ablations and transport diagnostics further characterize the contributions of individual components and the learned correspondence patterns.
Additional computational profiling in the appendix analyzes training and inference costs.

\subsection{Experimental Setup}

\paragraph{Datasets}
For long-term and few-shot forecasting, we evaluate CLAIR on six widely used multivariate benchmarks: Weather~\citep{Wu2021autoformer}, Electricity~\citep{Wu2021autoformer}, and four ETT datasets, namely ETTh1, ETTh2, ETTm1, and ETTm2~\citep{Zhou2021informer}. For short-term forecasting, we use the M4 benchmark~\citep{Makridakis2018m4}. These datasets cover heterogeneous temporal resolutions, sampling rates, and domain characteristics.

\paragraph{Baselines}
We compare CLAIR with representative LLM-based and non-LLM forecasting methods.
LLM-based baselines include Time-LLM~\citep{Jin2024TIMELLM}, One Fits All (OFA)~\citep{Zhou2023OneFitsAll}, S$^2$IP-LLM~\citep{Pan2024S2IP}, CALF~\citep{Liu2025CALF}, FSCA~\citep{Hu2025FSCA}, and SE-LLM~\citep{Liu2026SELLM}.
These methods cover patch-based adaptation, semantic and structural conditioning, temporal modeling, and cross-modal alignment.
Non-LLM baselines include DLinear~\citep{Zeng2023dlinear}, iTransformer~\citep{Liu2024itransformer}, and PatchTST~\citep{Nie2023patchtst}.

\paragraph{Experimental Protocol}
Following standard benchmarks, we fix the lookback window to $L=512$ and evaluate performance across four future horizons $H \in \{96, 192, 336, 720\}$. Forecasting accuracy on multivariate series is evaluated using Mean Squared Error (MSE) and Mean Absolute Error (MAE). For short-term forecasting on the M4 benchmark, we strictly follow the official frequency-specific protocol across diverse sampling granularities (Yearly, Quarterly, Monthly, Weekly, Daily, and Hourly), reporting Symmetric Mean Absolute Percentage Error (SMAPE), Mean Absolute Scaled Error (MASE), and Overall Weighted Average (OWA).

\subsection{Forecasting Performance}

\paragraph{Long-term Forecasting}
Table~\ref{tab:long-term-forcasting_results} summarizes multivariate long-term forecasting results averaged across prediction horizons.
CLAIR achieves the lowest MSE on Weather and all ETT benchmarks, with similarly strong MAE results.
This pattern supports the effectiveness of the framework across different temporal resolutions, while FSCA's stronger performance on Electricity indicates that the relative advantage remains dataset-dependent.

\paragraph{Short-term Forecasting}
Table~\ref{tab:short_term_forcasting_results} reports results on the M4 benchmark.
CLAIR combines the lowest MASE with near-best SMAPE, yielding the lowest OWA among the compared methods.
This combination indicates a favorable balance between scaled absolute and percentage errors under the frequency-specific protocol, supporting the applicability of the framework beyond multivariate long-term forecasting.

\paragraph{Few-shot Forecasting}
Tables~\ref{tab:fewshot10_results} and~\ref{tab:fewshot5_results} report horizon-averaged forecasting results using $10\%$ and $5\%$ of the training data, respectively.
CLAIR outperforms the evaluated LLM-based baselines on most datasets in the $10\%$ setting and across all datasets in the $5\%$ setting.
This pattern supports the effectiveness of the proposed adaptation framework when task-specific supervision is limited, showing that its competitive performance persists under more restrictive data budgets.

% Table: Patch correspondence, forecast consistency, and teacher stabilization
% Place in Section 4.3.
% Requires: \usepackage{booktabs}
\begin{table}[t]
    \centering
    \caption{
Comparison of patch correspondence, FC, and EMA on ETTh1 and ETTm1,
averaged over four prediction horizons.
}
    \label{tab:correspondence_fc}
    \scriptsize
    \setlength{\tabcolsep}{3.5pt}
    \renewcommand{\arraystretch}{1.05}
    \begin{tabular}{lcccc}
        \toprule
        & \multicolumn{2}{c}{ETTh1}
        & \multicolumn{2}{c}{ETTm1} \\
        \cmidrule(lr){2-3}
        \cmidrule(lr){4-5}
        Model Variant & MSE & MAE & MSE & MAE \\
        \midrule
        Supervised only
            & 0.4434 & 0.4581 & 0.3715 & 0.4014 \\
        FC only
            & 0.4332 & 0.4455 & 0.3671 & 0.3912 \\
        Index only
            & 0.4325 & 0.4443 & 0.3652 & 0.3883 \\
        PTA only
            & 0.4307 & 0.4426 & 0.3660 & 0.3895 \\
        \midrule
        Index + FC ($\mu=0$)
            & 0.4371 & 0.4463 & 0.3692 & 0.3947 \\
        PTA + FC ($\mu=0$)
            & 0.4357 & 0.4457 & 0.3681 & 0.3923 \\
        \midrule
        Index + FC (EMA)
            & 0.4183 & 0.4357 & 0.3585 & 0.3876 \\
        \textbf{PTA + FC (EMA, CLAIR)}
            & \textbf{0.4094} & \textbf{0.4287}
            & \textbf{0.3493} & \textbf{0.3820} \\
        \bottomrule
    \end{tabular}
\end{table}

\subsection{Patch Correspondence and Forecast Consistency}
\label{sec:correspondence_fc}

\paragraph{Interaction among patch correspondence, FC, and EMA}
We compare three representation-level alignment settings--no alignment,
index-wise alignment, and PTA--each with and without forecast-level
consistency (FC), while keeping the student architecture, input
perturbations, and supervised forecasting objective fixed.
Index-wise alignment uses the same normalized post-backbone
representations as PTA but matches patches at identical temporal
positions:
\begin{equation}
    \mathcal{L}_{\mathrm{Index}}
    =
    \frac{1}{UN}
    \sum_{u=1}^{U}
    \sum_{i=1}^{N}
    \left\|
        \tilde{h}_{s,i}^{(u)}
        -
        \operatorname{sg}\left(\tilde{h}_{t,i}^{(u)}\right)
    \right\|_2^2,
    \label{eq:index_alignment}
\end{equation}
where $U=BM$, $N$ is the number of patch tokens, and
$\operatorname{sg}(\cdot)$ denotes stop-gradient.
Teacher-guided variants use EMA with $\mu=0.9999$ unless otherwise
specified. $\mu=0$ denotes direct teacher synchronization without
temporal averaging.

Table~\ref{tab:correspondence_fc} shows that PTA and
index-wise alignment achieve comparable forecasting performance
without FC. When combined with FC, PTA achieves lower MSE and MAE
on both datasets, reducing horizon-averaged MSE by 2.1\% on ETTh1
and 2.6\% on ETTm1 relative to Index+FC.
These results suggest that flexible patch correspondence is more
effective when coupled with forecast-level consistency.
The EMA comparison further shows that this advantage depends on
teacher stabilization. With direct teacher synchronization
($\mu=0$), the relative MSE reductions of PTA+FC over Index+FC
decrease to approximately 0.3\% on both datasets.

\paragraph{Comparison with alternative correspondence strategies}
To further examine the predictive value of flexible correspondence,
we compare PTA+FC with alternative strategies on Weather and ETTh2
while keeping FC and the training settings fixed.
FC only excludes patch-level alignment, whereas Index+FC
enforces same-index correspondence.
Uniform+FC assigns equal correspondence weights regardless of
representation similarity, while NN+FC selects a single nearest
teacher patch for each student patch.
Local-OT+FC retains soft transport but restricts correspondence
to nearby temporal positions. In contrast, PTA+FC permits soft,
representation-dependent correspondence across all patch positions.

%==================== Patch Correspondence Strategies ====================%
\begin{table}[t!]
\centering
\caption{
Comparison of patch correspondence strategies on Weather and ETTh2
across four forecasting horizons.
}
\label{tab:matching_strategies}

\setlength{\tabcolsep}{2pt}
\renewcommand{\arraystretch}{1.05}

\resizebox{\columnwidth}{!}{%
    \begin{tabular}{lc*{6}{cc}}
    \toprule
    \multicolumn{2}{c}{}
    & \multicolumn{2}{c}{PTA+FC}
    & \multicolumn{2}{c}{Index+FC}
    & \multicolumn{2}{c}{Local-OT+FC}
    & \multicolumn{2}{c}{Uniform+FC}
    & \multicolumn{2}{c}{NN+FC}
    & \multicolumn{2}{c}{FC only} \\

    \cmidrule(lr){3-4}
    \cmidrule(lr){5-6}
    \cmidrule(lr){7-8}
    \cmidrule(lr){9-10}
    \cmidrule(lr){11-12}
    \cmidrule(lr){13-14}

    Dataset & $H$
    & MSE & MAE
    & MSE & MAE
    & MSE & MAE
    & MSE & MAE
    & MSE & MAE
    & MSE & MAE \\
    \midrule

    \multirow{4}{*}{Weather}
    & 96
    & \textbf{0.1498} & \textbf{0.2001}
    & 0.1508 & 0.2042
    & 0.1503 & 0.2038
    & 0.1507 & 0.2042
    & 0.1509 & 0.2046
    & 0.1517 & 0.2057 \\

    & 192
    & \textbf{0.1923} & \textbf{0.2417}
    & 0.1967 & 0.2454
    & 0.1940 & 0.2448
    & 0.1964 & 0.2451
    & 0.1976 & 0.2459
    & 0.1991 & 0.2470 \\

    & 336
    & \textbf{0.2469} & \textbf{0.2848}
    & 0.2513 & 0.2876
    & 0.2494 & 0.2861
    & 0.2515 & 0.2895
    & 0.2520 & 0.2881
    & 0.2527 & 0.2892 \\

    & 720
    & \textbf{0.3152} & \textbf{0.3301}
    & 0.3197 & 0.3356
    & 0.3184 & 0.3348
    & 0.3193 & 0.3353
    & 0.3204 & 0.3355
    & 0.3216 & 0.3383 \\

    \midrule

    \multirow{4}{*}{ETTh2}
    & 96
    & \textbf{0.2840} & \textbf{0.3432}
    & 0.2895 & 0.3494
    & 0.2867 & 0.3475
    & 0.2882 & 0.3486
    & 0.2901 & 0.3507
    & 0.2915 & 0.3513 \\

    & 192
    & \textbf{0.3530} & \textbf{0.3890}
    & 0.3610 & 0.3982
    & 0.3580 & 0.3956
    & 0.3586 & 0.3970
    & 0.3617 & 0.4008
    & 0.3643 & 0.4132 \\

    & 336
    & \textbf{0.3715} & \textbf{0.4099}
    & 0.3775 & 0.4155
    & 0.3741 & 0.4128
    & 0.3762 & 0.4138
    & 0.3790 & 0.4159
    & 0.3822 & 0.4178 \\

    & 720
    & \textbf{0.4171} & \textbf{0.4461}
    & 0.4253 & 0.4531
    & 0.4193 & 0.4501
    & 0.4210 & 0.4511
    & 0.4308 & 0.4591
    & 0.4329 & 0.4603 \\

    \bottomrule
    \end{tabular}%
}
\end{table}

Table~\ref{tab:matching_strategies} reports results on Weather and ETTh2 across four forecasting horizons. FC only yields the largest MSE across all horizons, while Index+FC, Uniform+FC, and NN+FC achieve lower errors but are consistently outperformed by Local-OT+FC. These results suggest that patch-level supervision is beneficial and that representation-dependent soft transport provides advantages over fixed, uniform, or hard matching strategies. Notably, PTA+FC further improves upon Local-OT+FC, achieving the lowest MSE and MAE across all horizons on both datasets. Its improvements over Index+FC extend the evidence for flexible correspondence beyond ETTh1 and ETTm1, while its advantage over Local-OT+FC supports unconstrained soft transport as an effective design choice when combined with FC.

% Table: Component ablation study
% Place in Section 4.4.
% Requires: \usepackage{booktabs}
\begin{table}[t]
    \centering
    \caption{Component ablation study evaluating the contributions of modules on ETTh1 and ETTm1 averaged over $H$.}
    \label{tab:component_ablation}
    \scriptsize
    \setlength{\tabcolsep}{3.5pt}
    \renewcommand{\arraystretch}{1.05}
    \begin{tabular}{lcccc}
        \toprule
        & \multicolumn{2}{c}{ETTh1}
        & \multicolumn{2}{c}{ETTm1} \\
        \cmidrule(lr){2-3}
        \cmidrule(lr){4-5}
        Model Variant & MSE & MAE & MSE & MAE \\
        \midrule
        w/o Gaussian Noise    & 0.4272 & 0.4409 & 0.3598 & 0.3876 \\
        w/o Soft Prompts      & 0.4119 & 0.4302 & 0.3516 & 0.3829 \\
        w/o Structural Anchor & 0.4129 & 0.4300 & 0.3528 & 0.3835 \\
        Frozen WPE+LN         & 0.4156 & 0.4333 & 0.3577 & 0.3854 \\
        \midrule
        \textbf{Full Model (CLAIR)}
            & \textbf{0.4094} & \textbf{0.4287}
            & \textbf{0.3493} & \textbf{0.3820} \\
        \bottomrule
    \end{tabular}
\end{table}

\subsection{Ablation Study}
\label{sec:ablation}

\begin{figure*}[t!]
  \centering
  \includegraphics[width=\textwidth]{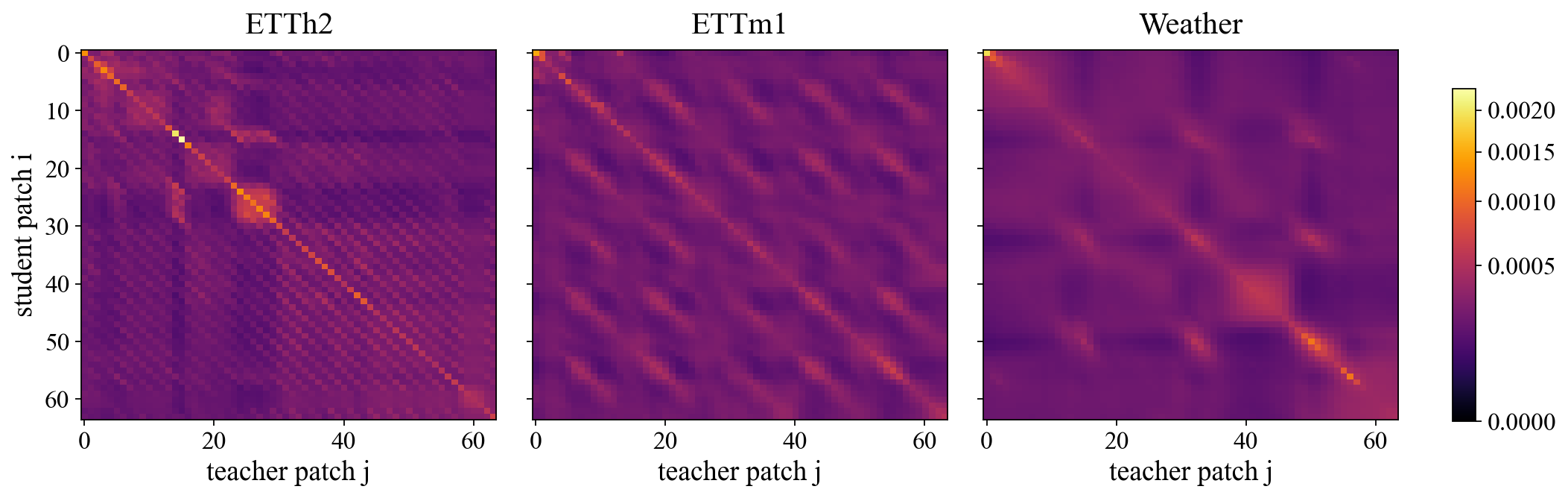}
  \caption{Batch-averaged unconstrained PTA transport plans $\pi$ at horizon $H=96$ on ETTh2 (left), ETTm1 (middle), and Weather (right). Rows and columns index student and teacher patches, respectively. }
  \label{fig:pta-mean-pi-h96}
\end{figure*}

\begin{figure*}[t]
  \centering
  \includegraphics[width=\textwidth]{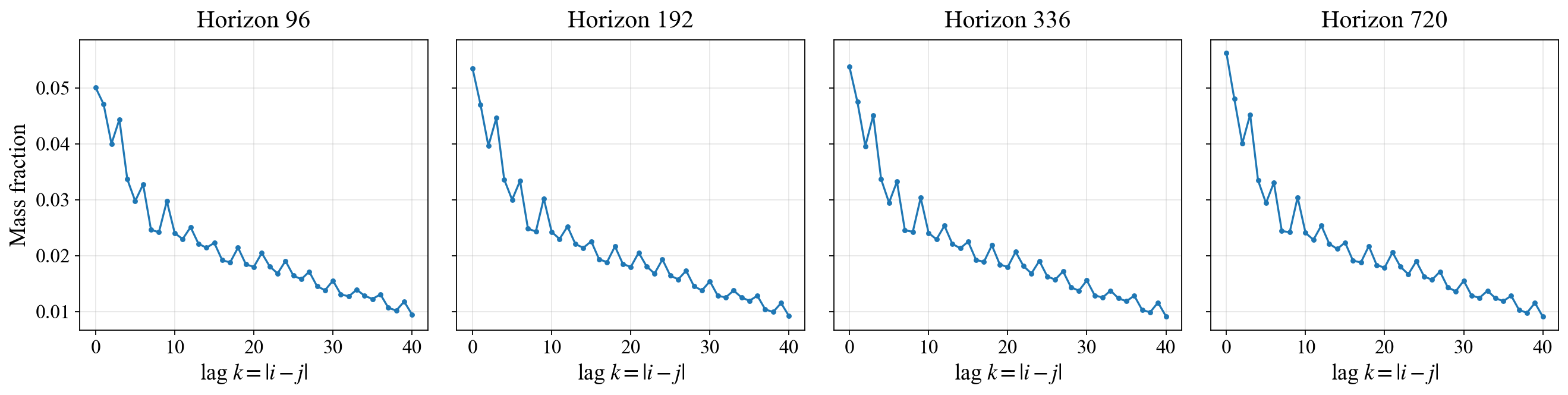}
  \caption{Transport mass fraction vs.\ temporal lag $k{=}|i{-}j|$ under the batch-averaged plan $\pi$ on ETTh2.}
  \label{fig:pta-etth2-lag-mass}
\end{figure*}

Table~\ref{tab:component_ablation} evaluates the contributions
of individual components on ETTh1 and ETTm1.
The full model achieves the lowest horizon-averaged MSE and MAE
among the evaluated variants on both datasets.
Removing student-side Gaussian noise produces the largest
performance degradation, while freezing GPT-2's positional
embeddings and layer-normalization parameters also increases
forecasting errors.
Removing the soft prompts or structural anchor leads to smaller
but consistent increases in error, supporting their contributions
to task-level and sequence-specific conditioning, respectively.

\begin{table}[t]
\centering
% \caption{Batch-averaged locality metrics of unconstrained PTA on ETTh2.}
\caption{Locality metrics of the mean unconstrained PTA plan $\bar{\Pi}$ on ETTh2, averaged over the first eight test windows (56 channel sequences) in evaluation mode.}
% \caption{Batch-averaged locality metrics of unconstrained PTA on ETTh2. Expected patch displacement is evaluated against the uniform baseline, alongside transport mass within near-diagonal bands ($|i{-}j| \le 1, 8$).}
\label{tab:pta-etth2-locality}
\begin{tabular}{lcccc}
\toprule
$H$ & $\mathbb{E}[|i{-}j|]$ & Unif.\ baseline & Band $|i{-}j|{\le}1$ & Band ${\le}8$ \\
\midrule
96  & 19.19 & 21.33 & 0.097 & 0.326 \\
192 & 18.94 & 21.33 & 0.100 & 0.331 \\
336 & 18.95 & 21.33 & 0.101 & 0.331 \\
720 & 18.85 & 21.33 & 0.104 & 0.334 \\
\bottomrule
\end{tabular}
\end{table}

\subsection{Temporal Structure of Unconstrained PTA}
\label{sec:pta_temporal_structure}

We examine whether unconstrained PTA exhibits temporal structure despite using neither an explicit temporal penalty nor a local-window constraint.
We first visualize mean transport plans on ETTh2, ETTm1, and Weather at $H{=}96$, and then analyze locality and lag--mass structure on ETTh2 across forecasting horizons $H\in\{96,192,336,720\}$.

\paragraph{Mean Transport Plans across Datasets}
At the fixed forecast horizon $H{=}96$, we average the student--teacher transport plans over the evaluated sequences separately for ETTh2, ETTm1, and Weather.
Figure~\ref{fig:pta-mean-pi-h96} shows the resulting dataset-specific mean transport plans, denoted by $\bar{\Pi}$.
Although the detailed patterns vary across datasets, the maps exhibit near-diagonal concentrations together with structured off-diagonal mass.
These patterns illustrate that PTA assigns mass to both local and non-local correspondences rather than restricting alignment to identical temporal positions.

\begin{figure*}[!t]
\centering

\begin{minipage}[t]{0.326\textwidth}
    \centering
    \includegraphics[width=\linewidth]{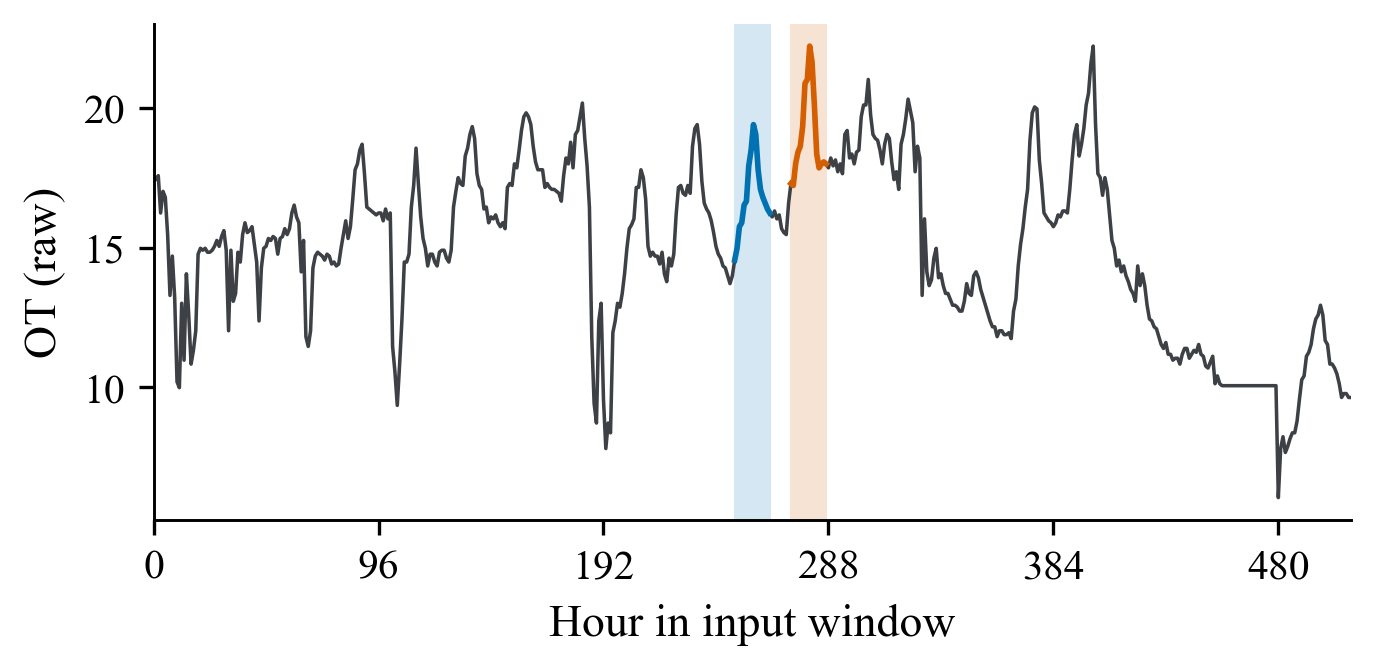}
    \par\smallskip
    {\small (a) Input window}
\end{minipage}
\hfill
\begin{minipage}[t]{0.326\textwidth}
    \centering
    \includegraphics[width=\linewidth]{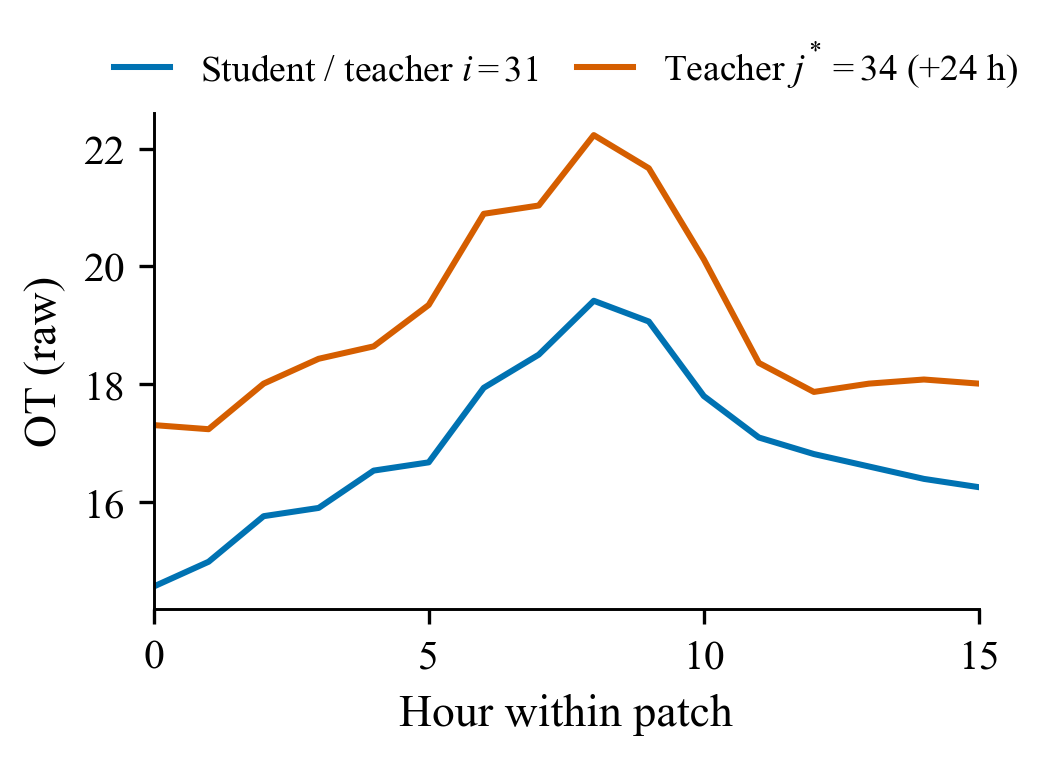}
    \par\smallskip
    {\small (b) Raw patch comparison}
\end{minipage}
\hfill
\begin{minipage}[t]{0.326\textwidth}
    \centering
    \includegraphics[width=\linewidth]{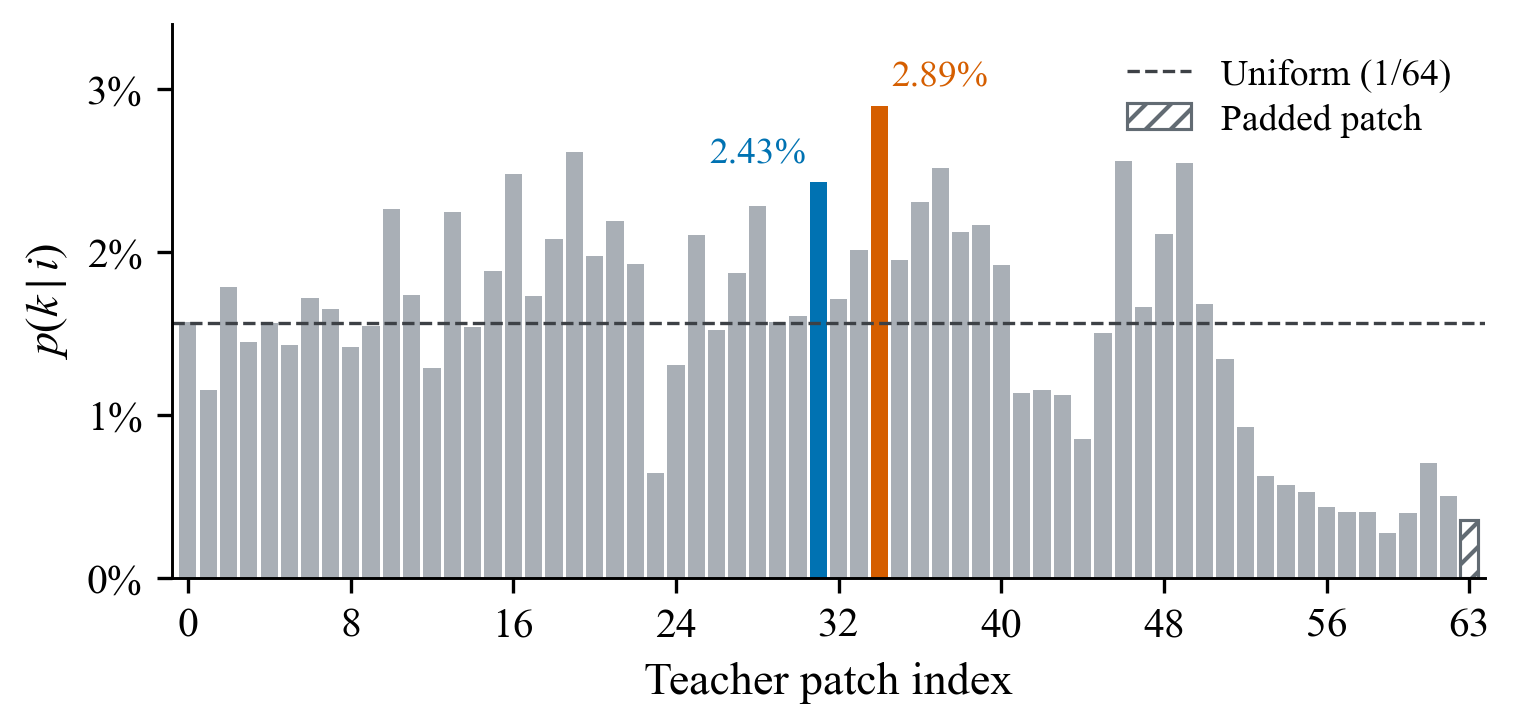}
    \par\smallskip
    {\small (c) Correspondence weights}
\end{minipage}

\caption{
Patch correspondence on the ETTh1 oil-temperature (OT) channel.
(a) A 512-hour input window with highlighted patches.
(b) Raw observations of student patch $i=31$ and
teacher patch $j^\star=34$.
(c) Row-normalized transport weights $p(j\mid i=31)$.
}
\label{fig:patch_correspondence_case}
\end{figure*}

\paragraph{Quantitative Locality on ETTh2}
We focus on ETTh2 for a detailed analysis of the near-diagonal concentration and repeated off-diagonal bands visible in Figure~\ref{fig:pta-mean-pi-h96}.
We first quantify locality across forecasting horizons using the expected patch-index displacement,
$\mathbb{E}_{\bar{\Pi}}[|i-j|]=\sum_{i,j}\bar{\Pi}_{ij}|i-j|$,
and the transport mass within near-diagonal bands $|i-j|\le b$.
Table~\ref{tab:pta-etth2-locality} shows that the expected displacement remains below the uniform-matching baseline across the evaluated horizons.
At the same time, the narrow band $|i-j|\le1$ accounts for only a small fraction of the total transport mass, and substantial mass remains outside the broader band $|i-j|\le8$.
Together, these results indicate a modest preference for nearby patches relative to uniform matching, while retaining substantial non-local correspondence.
The similar locality statistics across horizons suggest that this balance extends beyond the single forecast horizon visualized in Figure~\ref{fig:pta-mean-pi-h96}.

\paragraph{Lag--Mass Structure on ETTh2}
To characterize the temporal organization of the averaged transport plans, we aggregate transport mass by absolute patch-index lag:
$M(k)=\sum_{|i-j|=k}\bar{\Pi}_{ij}$. Note that $M(0)$  is the diagonal mass of the mean plan.
Figure~\ref{fig:pta-etth2-lag-mass} shows an overall decreasing trend with recurring peaks at approximately three-patch intervals across the evaluated horizons.
The overall decline should be interpreted cautiously because the number of available patch pairs also decreases at larger lags.
Under the hourly sampling rate of ETTh2 and patch stride $S{=}8$, a three-patch separation corresponds to $24$ hours.
The recurring peaks are therefore consistent with daily-scale structure in the averaged transport patterns.
These diagnostics characterize the temporal organization of PTA, complementing the controlled comparisons that evaluate the predictive effects of correspondence and forecast-level consistency.

\subsection{Qualitative Analysis of Patch Correspondence}
\label{sec:qualitative_correspondence}

To examine learned patch correspondence at the instance level,
we analyze a 512-hour input window from the oil-temperature
(OT) channel of ETTh1.
Figure~\ref{fig:patch_correspondence_case} shows the input
sequence, two selected patches, and the correspondence
weights for student patch $i=31$.
The transport plan is computed from the contextualized
representations of the trained student and EMA teacher,
using the same unperturbed input in evaluation mode.
Figure~\ref{fig:patch_correspondence_case}(a) marks
the input segments corresponding to student patch $i=31$
and teacher patch $j^\star=34$, whose starting positions
are 24 hours apart.
As shown in Figure~\ref{fig:patch_correspondence_case}(b),
the two non-overlapping patches exhibit similar rising
and falling patterns despite differences in their absolute levels.
Figure~\ref{fig:patch_correspondence_case}(c) shows
the row-normalized transport weights for student patch $i=31$.
The teacher patch at $j^\star=34$ receives a weight of
2.89\%, exceeding both the same-position weight of
2.43\% and the uniform reference of $1/64$.
Importantly, these weights are determined from contextualized
representations rather than direct similarity between
raw observations.
This example illustrates that PTA can assign greater
correspondence weight to a teacher representation at a
different temporal position than to its same-index counterpart,
while preserving the original patch order used for forecasting.

\section{Conclusion}

This paper presents CLAIR, an asymmetric teacher-guided
reprogramming framework that couples flexible PTA with FC for LLM-based time-series forecasting. Under a clean-view EMA teacher, PTA establishes soft
correspondences between contextualized patch representations,
while FC encourages agreement at corresponding future steps
without altering the original patch order used for forecasting.
Controlled comparisons show that PTA and index-wise alignment
perform comparably without FC, whereas PTA achieves lower
forecasting errors when combined with FC, particularly under
EMA supervision.
Additional comparisons with alternative correspondence
strategies further support the effectiveness of flexible
patch alignment, while component ablations highlight the
contributions of input perturbation and conditioning.
CLAIR achieves the lowest horizon-averaged MSE on five of
six long-term benchmarks and the lowest aggregate OWA on M4
among the evaluated methods, with competitive few-shot
performance. Transport diagnostics reveal both local and non-local
correspondence patterns, including recurring lag structures
consistent with daily-scale behavior on ETTh2.
Together, these findings support jointly considering patch
correspondence, forecast consistency, and teacher stabilization.
Although the EMA teacher and Sinkhorn transport introduce
training-time overhead, inference requires only the student.

% \onecolumn
\appendix

\section{Detailed Experimental Setup}
\label{app:setup}

\subsection{Implementation Details}
\label{app:impl}

All baselines are evaluated within a common experimental pipeline
built on Time-Series-Library~\citep{wu2023timesnet}.
Non-LLM baselines follow the library's official implementations,
whereas LLM-based methods use their publicly released codebases,
with input lengths, prediction horizons, and evaluation metrics
matched under the same protocol. Experiments are run on NVIDIA L40S GPUs and an
AMD EPYC 9654 CPU (96 cores).
Each configuration is repeated three times,
and we report the averaged results.

To reduce backbone-related differences, all LLM-based methods,
including Time-LLM, OFA, S$^2$IP-LLM, CALF, FSCA, SE-LLM,
and CLAIR, use the first six Transformer layers of pretrained
GPT-2 small~\citep{radford2019language}.
Time-LLM, OFA, S$^2$IP-LLM, CALF, and CLAIR use a patch length
of $P=16$ and stride $S=8$.
For FSCA, we adjust the patch length to $16$ and stride to $8$
in both long-term and few-shot forecasting to match this
common patching configuration.
For SE-LLM, we use a $512$-step input during both training
and evaluation, with $64$-step tokens and a $448$-step overlap
between the input and the shifted training target.

\subsection{Baselines}
\label{app:baselines}

\begin{itemize}
    \item \textbf{DLinear}~\citep{Zeng2023dlinear}: a linear forecasting model that decomposes series into trend and seasonal components.
    \item \textbf{PatchTST}~\citep{Nie2023patchtst}: a Transformer that tokenizes series into overlapping patches and applies temporal self-attention.
    \item \textbf{iTransformer}~\citep{Liu2024itransformer}: a Transformer variant that applies inverted attention across variables.
    \item \textbf{Time-LLM}~\citep{Jin2024TIMELLM}: reprograms time-series patches through text prototypes and Prompt-as-Prefix conditioning with a frozen LLM.
    \item \textbf{One Fits All (OFA)}~\citep{Zhou2023OneFitsAll}: unified LM adaptation by projecting patched continuous signals into soft tokens.
    \item \textbf{S$^2$IP-LLM}~\citep{Pan2024S2IP}: semantic-space-informed prompt learning with LLM embedding anchors.
    \item \textbf{CALF}~\citep{Liu2025CALF}: cross-modal fine-tuning that aligns numerical and language representations.
    \item \textbf{FSCA}~\citep{Hu2025FSCA}: context-level alignment of time-series and language inputs using dual-scale GNNs and few-shot prompting.
    \item \textbf{SE-LLM}~\citep{Liu2026SELLM}: semantic enhancement of LLM inputs with temporal patterns and short- and long-term dependency modeling.
\end{itemize}

\subsection{Evaluation Metrics}
\label{app:metrics}

For long-term and few-shot forecasting, we report MSE and MAE.
For short-term forecasting on M4, we report SMAPE, MASE,
and OWA~\citep{oreshkin2019nbeats},
following the official M4 protocol:

\begin{equation*}
\resizebox{\columnwidth}{!}{$\displaystyle
\begin{alignedat}{2}
\mathrm{MSE}
&= \frac{1}{H}\sum_{h=1}^{H}(Y_h-\hat Y_h)^2,
&\quad \mathrm{MAE}
&= \frac{1}{H}\sum_{h=1}^{H}|Y_h-\hat Y_h|,
\\[8pt]
\mathrm{SMAPE}
&= \frac{200}{H}\sum_{h=1}^{H}
   \frac{|Y_h-\hat Y_h|}{|Y_h|+|\hat Y_h|},
&\quad \mathrm{MAPE}
&= \frac{100}{H}\sum_{h=1}^{H}
   \frac{|Y_h-\hat Y_h|}{|Y_h|},
\\[8pt]
\mathrm{MASE}
&= \frac{\frac{1}{H}\sum_{h=1}^{H}|Y_h-\hat Y_h|}
        {\frac{1}{T-s}\sum_{t=s+1}^{T}|X_t-X_{t-s}|},
&\quad \mathrm{OWA}
&= \frac{1}{2}\left(
   \frac{\mathrm{SMAPE}}{\mathrm{SMAPE}_{\text{Na\"ive2}}}
   + \frac{\mathrm{MASE}}{\mathrm{MASE}_{\text{Na\"ive2}}}
   \right).
\end{alignedat}
$}
\end{equation*}

Here, $H$ is the prediction horizon, $T$ is the length of the
in-sample series, and $s$ is the seasonal periodicity.

\begin{table}[t]
\centering
\caption{Dataset statistics for long-term and short-term forecasting. Dataset size is reported as (training, validation, testing).}
\label{tab:app-dataset-stats}
\resizebox{\linewidth}{!}{%
\begin{tabular}{llcccc}
\toprule
Tasks & Datasets & Dim. & Forecast Horizon & Dataset Size & Frequency \\
\midrule
\multirow{6}{*}{Long-term}
& Weather & 21 & $\{96,192,336,720\}$ & (36792, 5271, 10540) & 10\,min \\
& ETTh1 & 7 & $\{96,192,336,720\}$ & (8545, 2881, 2881) & 1\,hour \\
& ETTh2 & 7 & $\{96,192,336,720\}$ & (8545, 2881, 2881) & 1\,hour \\
& ETTm1 & 7 & $\{96,192,336,720\}$ & (34465, 11521, 11521) & 15\,min \\
& ETTm2 & 7 & $\{96,192,336,720\}$ & (34465, 11521, 11521) & 15\,min \\
& Electricity & 321 & $\{96,192,336,720\}$ & (18317, 2633, 5261) & 1\,hour \\
\midrule
\multirow{6}{*}{Short-term}
& M4-Yearly & 1 & 6 & (23000, 0, 23000) & Yearly \\
& M4-Quarterly & 1 & 8 & (24000, 0, 24000) & Quarterly \\
& M4-Monthly & 1 & 18 & (48000, 0, 48000) & Monthly \\
& M4-Weekly & 1 & 13 & (359, 0, 359) & Weekly \\
& M4-Daily & 1 & 14 & (4227, 0, 4227) & Daily \\
& M4-Hourly & 1 & 48 & (414, 0, 414) & Hourly \\
\bottomrule
\end{tabular}%
}
\end{table}

\begin{table*}[t]
\centering
\caption{Multivariate long-term forecasting results (MSE / MAE) under full training (100\%).}
\label{tab:app-longterm-full}

\resizebox{\textwidth}{!}{%
\begin{tabular}{ll*{10}{cc}}
\toprule
& & \multicolumn{2}{c}{CLAIR}
  & \multicolumn{2}{c}{Time-LLM}
  & \multicolumn{2}{c}{S$^2$IP-LLM}
  & \multicolumn{2}{c}{OFA}
  & \multicolumn{2}{c}{CALF}
  & \multicolumn{2}{c}{FSCA}
  & \multicolumn{2}{c}{SE-LLM}
  & \multicolumn{2}{c}{DLinear}
  & \multicolumn{2}{c}{iTransformer}
  & \multicolumn{2}{c}{PatchTST} \\
\cmidrule(lr){3-4}
\cmidrule(lr){5-6}
\cmidrule(lr){7-8}
\cmidrule(lr){9-10}
\cmidrule(lr){11-12}
\cmidrule(lr){13-14}
\cmidrule(lr){15-16}
\cmidrule(lr){17-18}
\cmidrule(lr){19-20}
\cmidrule(lr){21-22}
Dataset & $H$
& MSE & MAE
& MSE & MAE
& MSE & MAE
& MSE & MAE
& MSE & MAE
& MSE & MAE
& MSE & MAE
& MSE & MAE
& MSE & MAE
& MSE & MAE \\
\midrule

\multirow{4}{*}{Weather}
& 96  & \underline{0.1498} & \textbf{0.2001}
      & 0.1663 & 0.2200
      & 0.1506 & \underline{0.2011}
      & \underline{0.1498} & 0.2020
      & 0.1571 & 0.2071
      & \textbf{0.1492} & 0.2025
      & 0.1521 & 0.2013
      & 0.1704 & 0.2298
      & 0.1708 & 0.2212
      & 0.1551 & 0.2123 \\
& 192 & \textbf{0.1923} & \textbf{0.2417}
      & 0.2024 & 0.2501
      & 0.1941 & 0.2426
      & 0.1944 & 0.2429
      & 0.2036 & 0.2494
      & \underline{0.1935} & \underline{0.2422}
      & 0.1977 & 0.2448
      & 0.2140 & 0.2719
      & 0.2118 & 0.2564
      & 0.2122 & 0.2636 \\
& 336 & \underline{0.2469} & \underline{0.2848}
      & 0.2500 & 0.2855
      & \textbf{0.2450} & \textbf{0.2810}
      & 0.2482 & \underline{0.2848}
      & 0.2523 & 0.2873
      & 0.2474 & 0.2856
      & 0.2505 & 0.2854
      & 0.2596 & 0.3112
      & 0.2604 & 0.2933
      & 0.2486 & 0.2857 \\
& 720 & \textbf{0.3152} & \textbf{0.3301}
      & 0.3197 & \underline{0.3338}
      & 0.3237 & 0.3383
      & 0.3216 & 0.3369
      & 0.3322 & 0.3437
      & 0.3221 & 0.3407
      & 0.3251 & 0.3387
      & 0.3225 & 0.3634
      & 0.3341 & 0.3391
      & \underline{0.3192} & 0.3378 \\
\midrule

\multirow{4}{*}{ETTh1}
& 96  & \underline{0.3695} & \underline{0.3982}
      & 0.4136 & 0.4292
      & 0.3711 & 0.4007
      & 0.3833 & 0.4049
      & 0.3863 & 0.4135
      & 0.3755 & 0.3997
      & 0.3783 & 0.4058
      & \textbf{0.3682} & \textbf{0.3981}
      & 0.3981 & 0.4227
      & 0.3834 & 0.4122 \\
& 192 & \textbf{0.4018} & \textbf{0.4192}
      & 0.4240 & 0.4337
      & 0.4026 & 0.4217
      & 0.4201 & 0.4273
      & 0.4176 & 0.4323
      & 0.4119 & 0.4217
      & 0.4264 & 0.4369
      & \underline{0.4025} & \underline{0.4205}
      & 0.4306 & 0.4450
      & 0.4696 & 0.4675 \\
& 336 & \textbf{0.4167} & \textbf{0.4300}
      & 0.4503 & 0.4583
      & \underline{0.4229} & 0.4442
      & 0.4356 & \underline{0.4375}
      & 0.4315 & 0.4432
      & 0.4324 & 0.4397
      & 0.4378 & 0.4387
      & 0.4311 & 0.4425
      & 0.4429 & 0.4560
      & 0.4712 & 0.4687 \\
& 720 & \textbf{0.4497} & \textbf{0.4673}
      & 0.4757 & 0.4854
      & 0.4649 & 0.4719
      & 0.4670 & 0.4726
      & 0.4769 & 0.4844
      & \underline{0.4521} & 0.4717
      & 0.4606 & \underline{0.4703}
      & 0.4718 & 0.4932
      & 0.5224 & 0.5212
      & 0.5158 & 0.5089 \\
\midrule

\multirow{4}{*}{ETTh2}
& 96  & \textbf{0.2840} & \textbf{0.3432}
      & 0.3068 & 0.3627
      & \underline{0.2856} & 0.3456
      & 0.2912 & 0.3458
      & 0.2864 & 0.3462
      & 0.2886 & \underline{0.3448}
      & 0.3086 & 0.3567
      & 0.3025 & 0.3676
      & 0.3038 & 0.3594
      & 0.3154 & 0.3715 \\
& 192 & \textbf{0.3530} & \textbf{0.3890}
      & 0.3700 & 0.3979
      & 0.3597 & 0.3977
      & 0.3656 & 0.3950
      & 0.3734 & 0.4001
      & \underline{0.3571} & \underline{0.3922}
      & 0.3818 & 0.4051
      & 0.3941 & 0.4261
      & 0.3755 & 0.4049
      & 0.3947 & 0.4234 \\
& 336 & \underline{0.3715} & \textbf{0.4099}
      & 0.3804 & 0.4143
      & \textbf{0.3662} & 0.4121
      & 0.4032 & 0.4284
      & 0.3867 & 0.4158
      & 0.3797 & \underline{0.4109}
      & 0.4115 & 0.4298
      & 0.4900 & 0.4895
      & 0.4182 & 0.4345
      & 0.4166 & 0.4415 \\
& 720 & \underline{0.4171} & \underline{0.4461}
      & 0.4173 & 0.4473
      & 0.4381 & 0.4654
      & 0.4369 & 0.4567
      & 0.4199 & \textbf{0.4458}
      & \textbf{0.4134} & 0.4493
      & 0.4274 & 0.4535
      & 0.8112 & 0.6381
      & 0.4410 & 0.4628
      & 0.4876 & 0.4855 \\
\midrule

\multirow{4}{*}{ETTm1}
& 96  & \textbf{0.2902} & \textbf{0.3467}
      & 0.3303 & 0.3685
      & \underline{0.2928} & 0.3496
      & 0.2996 & 0.3547
      & 0.3006 & 0.3503
      & 0.3186 & 0.3714
      & 0.3067 & 0.3545
      & 0.3037 & \underline{0.3471}
      & 0.3104 & 0.3639
      & 0.3135 & 0.3625 \\
& 192 & \underline{0.3330} & \underline{0.3715}
      & 0.3581 & 0.3843
      & \textbf{0.3324} & 0.3722
      & 0.3395 & 0.3802
      & 0.3370 & 0.3761
      & 0.3482 & 0.3771
      & 0.3463 & 0.3819
      & 0.3355 & \textbf{0.3662}
      & 0.3481 & 0.3874
      & 0.3640 & 0.3981 \\
& 336 & \textbf{0.3616} & \textbf{0.3905}
      & 0.3869 & 0.3993
      & \underline{0.3625} & 0.3919
      & 0.3681 & 0.3932
      & 0.3682 & 0.3946
      & 0.3675 & 0.3992
      & 0.3817 & 0.4046
      & 0.3714 & \underline{0.3909}
      & 0.3797 & 0.4045
      & 0.3918 & 0.4135 \\
& 720 & \textbf{0.4125} & \textbf{0.4192}
      & 0.4407 & 0.4344
      & \underline{0.4160} & \underline{0.4203}
      & 0.4312 & 0.4315
      & 0.4262 & 0.4268
      & 0.4353 & 0.4315
      & 0.4390 & 0.4381
      & 0.4236 & 0.4223
      & 0.4375 & 0.4396
      & 0.4388 & 0.4449 \\
\midrule

\multirow{4}{*}{ETTm2}
& 96  & \underline{0.1671} & \underline{0.2603}
      & 0.1809 & 0.2756
      & \textbf{0.1665} & \textbf{0.2563}
      & 0.1728 & 0.2662
      & 0.1721 & 0.2615
      & 0.1675 & 0.2606
      & 0.1757 & 0.2629
      & 0.1676 & 0.2633
      & 0.1801 & 0.2718
      & 0.1768 & 0.2693 \\
& 192 & \textbf{0.2230} & \textbf{0.2991}
      & 0.2310 & 0.3069
      & \underline{0.2247} & 0.3040
      & 0.2334 & 0.3074
      & 0.2294 & \underline{0.2993}
      & 0.2301 & 0.3003
      & 0.2404 & 0.3064
      & 0.2257 & 0.3056
      & 0.2421 & 0.3132
      & 0.2414 & 0.3107 \\
& 336 & \textbf{0.2740} & \textbf{0.3330}
      & 0.2795 & 0.3384
      & 0.2862 & 0.3426
      & 0.2868 & 0.3445
      & 0.2829 & \underline{0.3337}
      & \underline{0.2752} & 0.3359
      & 0.2992 & 0.3401
      & 0.2987 & 0.3618
      & 0.2920 & 0.3452
      & 0.2928 & 0.3460 \\
& 720 & \underline{0.3609} & \underline{0.3877}
      & 0.3682 & 0.3882
      & 0.3622 & 0.3916
      & 0.3721 & 0.3969
      & 0.3669 & \textbf{0.3867}
      & \textbf{0.3594} & 0.3881
      & 0.3907 & 0.4025
      & 0.4150 & 0.4380
      & 0.3728 & 0.3944
      & 0.4079 & 0.4152 \\
\midrule

\multirow{4}{*}{Electricity}
& 96
    & \underline{0.1337} & \underline{0.2290}
    & 0.1463 & 0.2502
    & 0.1378 & 0.2376
    & 0.1373 & 0.2367
    & 0.1373 & 0.2323
    & \textbf{0.1297} & \textbf{0.2251}
    & 0.1367 & 0.2330
    & 0.1406 & 0.2410
    & 0.1339 & 0.2294
    & 0.1369 & 0.2408 \\

& 192
    & \underline{0.1511} & \underline{0.2466}
    & 0.1587 & 0.2639
    & 0.1567 & 0.2557
    & 0.1534 & 0.2506
    & 0.1571 & 0.2528
    & \textbf{0.1506} & \textbf{0.2437}
    & 0.1536 & 0.2495
    & 0.1544 & 0.2542
    & 0.1566 & 0.2509
    & 0.1533 & 0.2548 \\

& 336
    & \underline{0.1681} & \textbf{0.2633}
    & 0.1737 & 0.2720
    & 0.1710 & 0.2693
    & 0.1686 & 0.2668
    & \textbf{0.1677} & \underline{0.2648}
    & 0.1689 & 0.2664
    & 0.1698 & 0.2657
    & 0.1693 & 0.2709
    & 0.1691 & 0.2652
    & 0.1691 & 0.2699 \\

& 720
    & \underline{0.2027} & 0.2945
    & 0.2089 & 0.3017
    & 0.2094 & 0.3013
    & 0.2061 & 0.2979
    & 0.2036 & 0.2946
    & 0.2047 & \underline{0.2942}
    & 0.2151 & 0.3026
    & 0.2040 & 0.3037
    & \textbf{0.1963} & \textbf{0.2882}
    & 0.2058 & 0.3001 \\

\bottomrule
\end{tabular}%
}
\end{table*}
\begin{table*}[t!]
\centering
\caption{Few-shot forecasting results with $10\%$ training data (MSE / MAE).}
\label{tab:app-fewshot-10-all}

\resizebox{\textwidth}{!}{%
\begin{tabular}{ll*{10}{cc}}
\toprule
& & \multicolumn{2}{c}{CLAIR}
  & \multicolumn{2}{c}{Time-LLM}
  & \multicolumn{2}{c}{S$^2$IP-LLM}
  & \multicolumn{2}{c}{OFA}
  & \multicolumn{2}{c}{CALF}
  & \multicolumn{2}{c}{FSCA}
  & \multicolumn{2}{c}{SE-LLM}
  & \multicolumn{2}{c}{DLinear}
  & \multicolumn{2}{c}{iTransformer}
  & \multicolumn{2}{c}{PatchTST} \\
\cmidrule(lr){3-4}
\cmidrule(lr){5-6}
\cmidrule(lr){7-8}
\cmidrule(lr){9-10}
\cmidrule(lr){11-12}
\cmidrule(lr){13-14}
\cmidrule(lr){15-16}
\cmidrule(lr){17-18}
\cmidrule(lr){19-20}
\cmidrule(lr){21-22}
Dataset & $H$
& MSE & MAE
& MSE & MAE
& MSE & MAE
& MSE & MAE
& MSE & MAE
& MSE & MAE
& MSE & MAE
& MSE & MAE
& MSE & MAE
& MSE & MAE \\
\midrule

\multirow{4}{*}{Weather}
& 96  & \underline{0.1626} & 0.2176
      & \textbf{0.1623} & \textbf{0.2144}
      & 0.1676 & 0.2226
      & 0.1650 & \underline{0.2155}
      & 0.1808 & 0.2319
      & 0.1684 & 0.2227
      & 0.1716 & 0.2189
      & 0.1750 & 0.2335
      & 0.1934 & 0.2446
      & 0.1739 & 0.2223 \\
& 192 & \textbf{0.2085} & \textbf{0.2577}
      & 0.2256 & 0.2731
      & \underline{0.2107} & 0.2604
      & 0.2155 & 0.2599
      & 0.2304 & 0.2747
      & 0.2134 & 0.2600
      & 0.2142 & \underline{0.2579}
      & 0.2147 & 0.2672
      & 0.2489 & 0.2893
      & 0.2327 & 0.2783 \\
& 336 & \textbf{0.2566} & \textbf{0.2915}
      & 0.2759 & 0.3118
      & 0.2584 & \underline{0.2917}
      & 0.2666 & 0.3000
      & 0.2779 & 0.3126
      & 0.2614 & 0.2981
      & 0.2699 & 0.3031
      & \underline{0.2579} & 0.3036
      & 0.2975 & 0.3288
      & 0.2704 & 0.3066 \\
& 720 & \underline{0.3212} & \textbf{0.3394}
      & 0.3325 & 0.3504
      & 0.3247 & \underline{0.3402}
      & 0.3248 & 0.3416
      & 0.3401 & 0.3555
      & 0.3267 & 0.3452
      & 0.3489 & 0.3576
      & \textbf{0.3202} & 0.3534
      & 0.4175 & 0.4027
      & 0.3346 & 0.3498 \\
\midrule

\multirow{4}{*}{ETTh1}
& 96  & \textbf{0.4503} & \textbf{0.4407}
      & 0.5204 & 0.4903
      & 0.5936 & 0.5230
      & 0.4582 & 0.4591
      & 0.5457 & 0.5153
      & 0.4669 & 0.4663
      & 0.4831 & 0.4600
      & \underline{0.4508} & \underline{0.4530}
      & 0.5952 & 0.5270
      & 0.5631 & 0.5158 \\
& 192 & \underline{0.5006} & \underline{0.4727}
      & 0.5790 & 0.5156
      & 0.6828 & 0.5388
      & 0.5363 & 0.4890
      & 0.5978 & 0.5425
      & 0.5225 & 0.4865
      & 0.5568 & 0.5020
      & \textbf{0.4808} & \textbf{0.4702}
      & 0.6509 & 0.5607
      & 0.7542 & 0.5990 \\
& 336 & \underline{0.5337} & \underline{0.4946}
      & 0.7554 & 0.5819
      & 0.7217 & 0.5437
      & 0.6292 & 0.5487
      & 0.7365 & 0.5843
      & 0.6684 & 0.5605
      & 0.6145 & 0.5379
      & \textbf{0.4994} & \textbf{0.4875}
      & 0.7080 & 0.5744
      & 0.7644 & 0.6183 \\
& 720 & 0.7508 & 0.6174
      & 0.8969 & 0.6461
      & 0.9745 & 0.6981
      & \underline{0.7410} & \underline{0.6083}
      & 0.9872 & 0.7020
      & 0.8769 & 0.6424
      & 0.9134 & 0.6707
      & \textbf{0.6924} & \textbf{0.5991}
      & 0.8565 & 0.6526
      & 0.8218 & 0.6532 \\
\midrule

\multirow{4}{*}{ETTh2}
& 96  & \textbf{0.3033} & \textbf{0.3636}
      & 0.3197 & 0.3791
      & 0.3454 & 0.3934
      & 0.3222 & 0.3788
      & \underline{0.3159} & 0.3771
      & 0.3365 & 0.3773
      & 0.3194 & \underline{0.3770}
      & 0.3301 & 0.3834
      & 0.3620 & 0.4029
      & 0.3908 & 0.4323 \\
& 192 & \textbf{0.3837} & \underline{0.4163}
      & 0.3955 & 0.4246
      & 0.4149 & 0.4382
      & 0.4235 & 0.4422
      & 0.4218 & 0.4343
      & 0.4212 & 0.4396
      & \underline{0.3888} & \textbf{0.4133}
      & 0.4205 & 0.4455
      & 0.4167 & 0.4366
      & 0.4677 & 0.4602 \\
& 336 & \textbf{0.4102} & \textbf{0.4430}
      & 0.4193 & 0.4468
      & 0.4215 & 0.4476
      & 0.4457 & 0.4599
      & 0.4304 & 0.4484
      & 0.4262 & 0.4482
      & \underline{0.4148} & \underline{0.4438}
      & 0.4389 & 0.4590
      & 0.4330 & 0.4536
      & 0.4713 & 0.4851 \\
& 720 & \textbf{0.4411} & \textbf{0.4676}
      & 0.4821 & 0.4827
      & 0.6771 & 0.5849
      & 0.5883 & 0.5504
      & 0.5148 & 0.5089
      & 0.4609 & \underline{0.4728}
      & \underline{0.4572} & 0.4746
      & 0.4651 & 0.4958
      & 0.4929 & 0.4959
      & 0.4924 & 0.4860 \\
\midrule

\multirow{4}{*}{ETTm1}
& 96  & \underline{0.3881} & \underline{0.3988}
      & 0.4331 & 0.4297
      & 0.4165 & 0.4231
      & 0.3940 & 0.4048
      & 0.4109 & 0.4235
      & 0.4393 & 0.4264
      & 0.5337 & 0.4735
      & \textbf{0.3615} & \textbf{0.3955}
      & 0.4635 & 0.4536
      & 0.4423 & 0.4297 \\
& 192 & \underline{0.4211} & \underline{0.4196}
      & 0.4492 & 0.4386
      & 0.4613 & 0.4495
      & 0.4242 & 0.4200
      & 0.4594 & 0.4463
      & 0.4579 & 0.4294
      & 0.6017 & 0.5145
      & \textbf{0.3862} & \textbf{0.4109}
      & 0.4874 & 0.4716
      & 0.5438 & 0.4773 \\
& 336 & \underline{0.4490} & \underline{0.4325}
      & 0.4762 & 0.4558
      & 0.4719 & 0.4555
      & 0.4591 & 0.4360
      & 0.5345 & 0.4955
      & 0.4607 & 0.4355
      & 0.6444 & 0.5172
      & \textbf{0.4182} & \textbf{0.4298}
      & 0.6144 & 0.5346
      & 0.6570 & 0.5242 \\
& 720 & \underline{0.5346} & \underline{0.4830}
      & 0.5388 & 0.4894
      & 0.5839 & 0.5117
      & 0.5621 & 0.4954
      & 0.5890 & 0.5230
      & 0.6115 & 0.5248
      & 0.7239 & 0.5557
      & \textbf{0.4864} & \textbf{0.4737}
      & 0.7168 & 0.5804
      & 0.7661 & 0.5819 \\
\midrule

\multirow{4}{*}{ETTm2}
& 96  & \textbf{0.1704} & \textbf{0.2606}
      & 0.2083 & 0.2978
      & 0.2082 & 0.2881
      & 0.1980 & 0.2780
      & \underline{0.1903} & \underline{0.2750}
      & 0.1980 & 0.2764
      & 0.2025 & 0.2820
      & 0.2068 & 0.2928
      & 0.2180 & 0.3011
      & 0.2126 & 0.2942 \\
& 192 & \textbf{0.2265} & \textbf{0.2997}
      & 0.2609 & 0.3296
      & 0.2770 & 0.3355
      & 0.2537 & 0.3128
      & 0.2550 & 0.3200
      & 0.2526 & \underline{0.3106}
      & \underline{0.2477} & 0.3174
      & 0.2649 & 0.3318
      & 0.2745 & 0.3355
      & 0.2813 & 0.3397 \\
& 336 & \textbf{0.2827} & \textbf{0.3369}
      & 0.3039 & 0.3542
      & 0.3209 & 0.3656
      & 0.2998 & \underline{0.3432}
      & 0.3150 & 0.3599
      & 0.3059 & 0.3559
      & \underline{0.2988} & 0.3490
      & 0.3174 & 0.3656
      & 0.3237 & 0.3676
      & 0.3314 & 0.3679 \\
& 720 & \textbf{0.3648} & \textbf{0.3896}
      & 0.4338 & 0.4245
      & 0.4247 & 0.4230
      & 0.4081 & 0.4091
      & 0.4173 & 0.4160
      & \underline{0.3883} & \underline{0.3934}
      & 0.4076 & 0.4192
      & 0.3969 & 0.4126
      & 0.4424 & 0.4360
      & 0.4942 & 0.4565 \\
\midrule

\multirow{4}{*}{Electricity}
& 96  & \underline{0.1419} & \textbf{0.2409}
      & 0.1501 & 0.2522
      & 0.1438 & 0.2444
      & \textbf{0.1418} & 0.2414
      & 0.1495 & 0.2523
      & 0.1422 & \underline{0.2413}
      & 0.1509 & 0.2471
      & 0.1945 & 0.3060
      & 0.1505 & 0.2537
      & 0.1494 & 0.2532 \\
& 192 & 0.1611 & 0.2593
      & 0.1645 & 0.2675
      & \underline{0.1585} & 0.2568
      & \textbf{0.1580} & \underline{0.2554}
      & 0.1710 & 0.2729
      & 0.1587 & \textbf{0.2553}
      & 0.1680 & 0.2629
      & 0.2058 & 0.3159
      & 0.1730 & 0.2740
      & 0.1655 & 0.2680 \\
& 336 & 0.1821 & 0.2803
      & 0.1861 & 0.2875
      & \textbf{0.1773} & \underline{0.2745}
      & \underline{0.1774} & \textbf{0.2733}
      & 0.1944 & 0.2946
      & 0.1826 & 0.2765
      & 0.1923 & 0.2894
      & 0.2211 & 0.3306
      & 0.2043 & 0.3029
      & 0.1842 & 0.2867 \\
& 720 & 0.2426 & 0.3307
      & 0.2432 & 0.3310
      & \textbf{0.2355} & \underline{0.3216}
      & \textbf{0.2355} & \textbf{0.3192}
      & 0.2577 & 0.3437
      & 0.2446 & 0.3316
      & 0.2454 & 0.3323
      & 0.2590 & 0.3625
      & 0.2506 & 0.3347
      & \underline{0.2400} & 0.3284 \\

\bottomrule
\end{tabular}%
}
\end{table*}
\begin{table*}[t!]
\centering
\caption{Few-shot forecasting results with $5\%$ training data (MSE / MAE).}
\label{tab:app-fewshot-5-full}

\resizebox{\textwidth}{!}{%
\begin{tabular}{ll*{10}{cc}}
\toprule
& & \multicolumn{2}{c}{CLAIR}
  & \multicolumn{2}{c}{Time-LLM}
  & \multicolumn{2}{c}{S$^2$IP-LLM}
  & \multicolumn{2}{c}{OFA}
  & \multicolumn{2}{c}{CALF}
  & \multicolumn{2}{c}{FSCA}
  & \multicolumn{2}{c}{SE-LLM}
  & \multicolumn{2}{c}{DLinear}
  & \multicolumn{2}{c}{iTransformer}
  & \multicolumn{2}{c}{PatchTST} \\
\cmidrule(lr){3-4}
\cmidrule(lr){5-6}
\cmidrule(lr){7-8}
\cmidrule(lr){9-10}
\cmidrule(lr){11-12}
\cmidrule(lr){13-14}
\cmidrule(lr){15-16}
\cmidrule(lr){17-18}
\cmidrule(lr){19-20}
\cmidrule(lr){21-22}
Dataset & $H$
& MSE & MAE
& MSE & MAE
& MSE & MAE
& MSE & MAE
& MSE & MAE
& MSE & MAE
& MSE & MAE
& MSE & MAE
& MSE & MAE
& MSE & MAE \\
\midrule

\multirow{4}{*}{Weather}
& 96  & \underline{0.1798} & 0.2381
      & \textbf{0.1775} & \textbf{0.2310}
      & 0.1884 & 0.2423
      & 0.1855 & 0.2349
      & 0.1870 & 0.2412
      & 0.1841 & \underline{0.2338}
      & 0.1917 & 0.2439
      & 0.1922 & 0.2598
      & 0.2088 & 0.2617
      & 0.1854 & 0.2390 \\
& 192 & \textbf{0.2290} & \textbf{0.2796}
      & 0.2617 & 0.2958
      & 0.2438 & 0.2969
      & 0.2377 & 0.2870
      & 0.2401 & 0.2867
      & \underline{0.2322} & \underline{0.2824}
      & 0.2389 & 0.2843
      & 0.2413 & 0.3019
      & 0.2569 & 0.3015
      & 0.2492 & 0.2917 \\
& 336 & \textbf{0.2896} & \underline{0.3284}
      & 0.2970 & \textbf{0.3268}
      & 0.3289 & 0.3531
      & 0.2921 & 0.3309
      & 0.2949 & 0.3288
      & \underline{0.2913} & 0.3302
      & 0.3043 & 0.3296
      & 0.2928 & 0.3418
      & 0.3138 & 0.3480
      & 0.3108 & 0.3437 \\
& 720 & \textbf{0.3582} & \textbf{0.3723}
      & 0.3748 & 0.3850
      & 0.4061 & 0.4053
      & 0.3728 & 0.3827
      & 0.3814 & 0.3883
      & \underline{0.3605} & \underline{0.3756}
      & 0.3629 & 0.3770
      & 0.3632 & 0.3945
      & 0.3875 & 0.3978
      & 0.3928 & 0.3975 \\
\midrule

\multirow{4}{*}{ETTh1}
& 96  & \underline{0.6033} & \underline{0.5157}
      & 0.7049 & 0.5613
      & 0.8031 & 0.5903
      & 0.6220 & 0.5297
      & 0.6659 & 0.5519
      & 0.6313 & 0.5496
      & 0.6403 & 0.5528
      & \textbf{0.5565} & \textbf{0.5057}
      & 0.7338 & 0.5789
      & 0.6462 & 0.5551 \\
& 192 & \underline{0.6689} & \textbf{0.5432}
      & 0.8215 & 0.5879
      & 0.8289 & 0.5942
      & 0.8224 & 0.6094
      & 0.9616 & 0.6758
      & 0.7869 & 0.6020
      & 0.8295 & 0.6274
      & \textbf{0.6305} & \underline{0.5550}
      & 0.8229 & 0.6188
      & 0.9925 & 0.7022 \\
& 336 & \underline{0.7436} & \underline{0.5939}
      & 0.9112 & 0.6264
      & 1.3839 & 0.8311
      & 1.0763 & 0.7180
      & 0.9742 & 0.6850
      & 0.8076 & 0.6144
      & 0.8558 & 0.6289
      & \textbf{0.6937} & \textbf{0.5828}
      & 0.8844 & 0.6493
      & 1.0182 & 0.7142 \\
& 720 & -- & -- & -- & -- & -- & -- & -- & -- & -- & --
      & -- & -- & -- & --
      & -- & -- & -- & -- & -- & -- \\
\midrule

\multirow{4}{*}{ETTh2}
& 96  & \textbf{0.3390} & 0.4061
      & 0.4671 & 0.4686
      & 0.4973 & 0.4827
      & 0.4204 & 0.4665
      & 0.3474 & \underline{0.4019}
      & 0.3437 & 0.4092
      & 0.3556 & 0.4058
      & 0.4128 & 0.4588
      & 0.3941 & 0.4262
      & \underline{0.3404} & \textbf{0.3917} \\
& 192 & \textbf{0.3768} & \textbf{0.4236}
      & 0.5475 & 0.5082
      & 0.5978 & 0.5380
      & 0.4533 & 0.4532
      & 0.4845 & 0.4843
      & \underline{0.3806} & \underline{0.4256}
      & 0.4796 & 0.4935
      & 0.4611 & 0.4943
      & 0.4413 & 0.4564
      & 0.4118 & 0.4262 \\
& 336 & \underline{0.4029} & \underline{0.4408}
      & 0.6252 & 0.5532
      & 0.7550 & 0.6154
      & 0.5354 & 0.5335
      & 0.4914 & 0.4912
      & \textbf{0.4018} & \textbf{0.4397}
      & 0.5448 & 0.5428
      & 0.5878 & 0.5739
      & 0.4586 & 0.4722
      & 0.4633 & 0.4643 \\
& 720 & -- & -- & -- & -- & -- & -- & -- & -- & -- & --
      & -- & -- & -- & --
      & -- & -- & -- & -- & -- & -- \\
\midrule

\multirow{4}{*}{ETTm1}
& 96  & \underline{0.3884} & \textbf{0.4006}
      & 0.3942 & 0.4110
      & 0.4083 & 0.4225
      & 0.3932 & 0.4093
      & 0.3986 & 0.4149
      & 0.4515 & 0.4315
      & 0.5534 & 0.4776
      & \textbf{0.3783} & \underline{0.4068}
      & 0.4227 & 0.4334
      & 0.4318 & 0.4325 \\
& 192 & \underline{0.4209} & \textbf{0.4232}
      & 0.4427 & 0.4397
      & 0.4387 & 0.4382
      & 0.4382 & 0.4341
      & 0.4377 & 0.4405
      & 0.4346 & 0.4286
      & 0.6009 & 0.4983
      & \textbf{0.4103} & \underline{0.4272}
      & 0.4906 & 0.4698
      & 0.5300 & 0.4912 \\
& 336 & \underline{0.4701} & \underline{0.4526}
      & 0.5577 & 0.5001
      & 0.5167 & 0.4887
      & 0.5048 & 0.4677
      & 0.4960 & 0.4750
      & 0.4849 & 0.4528
      & 0.6024 & 0.5013
      & \textbf{0.4455} & \textbf{0.4444}
      & 0.5697 & 0.5104
      & 0.5532 & 0.4977 \\
& 720 & \underline{0.5621} & \textbf{0.5042}
      & 0.6363 & 0.5406
      & 0.7168 & 0.5945
      & 0.6395 & 0.5389
      & 0.7646 & 0.6094
      & 0.6168 & 0.5389
      & 0.6676 & 0.5490
      & \textbf{0.5514} & \underline{0.5068}
      & 0.7707 & 0.6052
      & 0.7661 & 0.5975 \\
\midrule

\multirow{4}{*}{ETTm2}
& 96  & \underline{0.2024} & \underline{0.2847}
      & 0.2185 & 0.2962
      & 0.2205 & 0.2941
      & \textbf{0.2006} & \textbf{0.2816}
      & 0.2167 & 0.2963
      & 0.2123 & 0.2899
      & 0.2143 & 0.2965
      & 0.2204 & 0.3031
      & 0.2126 & 0.2987
      & 0.2353 & 0.3089 \\
& 192 & \textbf{0.2523} & \textbf{0.3142}
      & 0.2814 & 0.3423
      & 0.2785 & 0.3354
      & 0.2600 & 0.3201
      & 0.2865 & 0.3388
      & \underline{0.2576} & \underline{0.3155}
      & 0.2754 & 0.3282
      & 0.2789 & 0.3440
      & 0.2673 & 0.3340
      & 0.3113 & 0.3611 \\
& 336 & \textbf{0.3024} & \textbf{0.3481}
      & 0.3225 & 0.3606
      & 0.3277 & 0.3723
      & 0.3264 & 0.3607
      & 0.3462 & 0.3832
      & 0.3263 & 0.3630
      & \underline{0.3109} & \underline{0.3580}
      & 0.3263 & 0.3751
      & 0.3189 & 0.3665
      & 0.3655 & 0.3866 \\
& 720 & \textbf{0.4159} & \textbf{0.4151}
      & 0.4687 & 0.4479
      & 0.4527 & 0.4476
      & 0.4824 & 0.4442
      & \underline{0.4294} & 0.4209
      & \underline{0.4294} & \underline{0.4206}
      & 0.4324 & 0.4227
      & 0.4795 & 0.4824
      & 0.4352 & 0.4301
      & 0.5010 & 0.4581 \\
\midrule

\multirow{4}{*}{Electricity}
& 96  & \textbf{0.1443} & \textbf{0.2439}
      & 0.1515 & 0.2549
      & \underline{0.1455} & \underline{0.2452}
      & 0.1506 & 0.2538
      & 0.1657 & 0.2701
      & 0.1469 & 0.2461
      & 0.1592 & 0.2557
      & 0.2242 & 0.3338
      & 0.1578 & 0.2616
      & 0.1554 & 0.2602 \\
& 192 & \underline{0.1616} & \underline{0.2603}
      & 0.1708 & 0.2732
      & \textbf{0.1610} & \textbf{0.2596}
      & 0.1655 & 0.2662
      & 0.1926 & 0.2946
      & 0.1624 & 0.2611
      & 0.1757 & 0.2718
      & 0.2403 & 0.3471
      & 0.1843 & 0.2880
      & 0.1707 & 0.2731 \\
& 336 & 0.1831 & \underline{0.2799}
      & 0.1917 & 0.2912
      & \textbf{0.1818} & \textbf{0.2797}
      & 0.1865 & 0.2858
      & 0.2272 & 0.3248
      & \underline{0.1829} & 0.2818
      & 0.1989 & 0.2963
      & 0.2643 & 0.3666
      & 0.2104 & 0.3129
      & 0.1906 & 0.2914 \\
& 720 & \textbf{0.2333} & \textbf{0.3185}
      & 0.2418 & 0.3359
      & 0.2442 & 0.3359
      & 0.2376 & \underline{0.3257}
      & 0.3268 & 0.4030
      & \underline{0.2363} & \underline{0.3257}
      & 0.2528 & 0.3406
      & 0.4542 & 0.5266
      & 0.2882 & 0.3733
      & 0.2446 & 0.3356 \\

\bottomrule
\end{tabular}%
}
\end{table*}
\begin{table*}[t!]
\centering
\caption{Short-term forecasting results on the M4 dataset (SMAPE / MASE / OWA).}
\label{tab:short_term_forcasting_results_full}

\resizebox{\textwidth}{!}{
    % \begin{tabular}{c|l|cccccccccc}
    \begin{tabular}{clcccccccccc}
    \toprule
    % \multicolumn{2}{c|}{Models}
    \multicolumn{2}{c}{}
    & CLAIR
    & Time-LLM
    & S\textsuperscript{2}IP-LLM
    & OFA
    & CALF
    & FSCA
    & SE-LLM
    & DLinear
    & iTransformer
    & PatchTST \\
    \midrule

    \multirow{3}{*}{Yearly}
    & SMAPE
    & 13.764
    & 13.721
    & 13.899
    & 14.368
    & \underline{13.488}
    & \textbf{13.413}
    & 13.675
    & 14.358
    & 14.282
    & 13.664 \\

    & MASE
    & 3.161
    & \underline{3.036}
    & 3.093
    & 3.397
    & 3.043
    & \textbf{3.009}
    & 3.140
    & 3.128
    & 3.202
    & 3.086 \\

    & OWA
    & 0.819
    & 0.802
    & 0.814
    & 0.867
    & \underline{0.796}
    & \textbf{0.789}
    & 0.813
    & 0.833
    & 0.840
    & 0.806 \\
    \midrule

    \multirow{3}{*}{Quarterly}
    & SMAPE
    & \textbf{10.116}
    & 10.861
    & 10.545
    & 10.444
    & 10.697
    & 10.213
    & \underline{10.191}
    & 10.500
    & 10.740
    & 10.916 \\

    & MASE
    & \textbf{1.178}
    & 1.293
    & 1.239
    & 1.237
    & 1.233
    & 1.227
    & \underline{1.193}
    & 1.239
    & 1.280
    & 1.289 \\

    & OWA
    & \textbf{0.889}
    & 0.965
    & 0.931
    & 0.925
    & 0.935
    & 0.911
    & \underline{0.898}
    & 0.929
    & 0.954
    & 0.966 \\
    \midrule

    \multirow{3}{*}{Monthly}
    & SMAPE
    & \textbf{12.858}
    & 13.445
    & 13.122
    & \underline{12.871}
    & 13.068
    & 12.963
    & 13.121
    & 13.394
    & 15.062
    & 14.180 \\

    & MASE
    & \textbf{0.953}
    & 1.027
    & 0.987
    & \textbf{0.953}
    & \underline{0.968}
    & 0.987
    & 0.983
    & 1.007
    & 1.170
    & 1.127 \\

    & OWA
    & \textbf{0.894}
    & 0.949
    & 0.919
    & \textbf{0.894}
    & \underline{0.908}
    & 0.913
    & 0.917
    & 0.938
    & 1.072
    & 1.021 \\
    \midrule

    \multirow{3}{*}{Others}
    & SMAPE
    & \textbf{4.848}
    & 6.122
    & 5.763
    & 5.321
    & 5.011
    & \underline{4.893}
    & 5.124
    & 5.111
    & 5.342
    & 6.089 \\

    & MASE
    & \textbf{3.225}
    & 4.060
    & 4.023
    & 3.566
    & \underline{3.418}
    & 3.534
    & 3.508
    & 3.665
    & 3.613
    & 3.941 \\

    & OWA
    & \textbf{1.019}
    & 1.284
    & 1.241
    & 1.122
    & \underline{1.066}
    & 1.072
    & 1.092
    & 1.116
    & 1.132
    & 1.262 \\

    \bottomrule
    \end{tabular}
}
\end{table*}

\subsection{Dataset Descriptions}
\label{app:datasets}

Table~\ref{tab:app-dataset-stats} summarizes dataset statistics.
For long-term and few-shot forecasting we use Weather and Electricity~\citep{Wu2021autoformer} and four ETT datasets (ETTh1, ETTh2, ETTm1, ETTm2)~\citep{Zhou2021informer}, with input length $L{=}512$ and horizons $H\in\{96,192,336,720\}$.
For short-term forecasting we use the M4 benchmark~\citep{Makridakis2018m4}.

\subsection{Teacher-guided Hyperparameter Configuration}
\label{app:hyper}

\textbf{Default shared settings.}
We use $n_p=4$ soft prompt tokens, $K=3$ temporal scales in the structural anchor,
EMA momentum $\mu=0.9999$, warm-up $E_{\mathrm{warm}}=5$, student Gaussian noise $\sigma=0.01$,
student dropout 0.05, and Sinkhorn parameters $\epsilon_{\mathrm{OT}}=0.1$ with 20 iterations.

\textbf{Loss weights.}
We search both the PTA and FC loss weights, $\lambda_{\mathrm{PTA}}$ and $\lambda_{\mathrm{FC}}$, over $\{0.05,0.10,0.15,0.20\}$.
The PTA weight is fixed at $\lambda_{\mathrm{PTA}}{=}0.10$ across all six long-term forecasting datasets.
The FC weight is set to $\lambda_{\mathrm{FC}}{=}0.10$ for ETTh1 and $\lambda_{\mathrm{FC}}{=}0.05$ for ETTh2, ETTm1, ETTm2, Weather, and Electricity.

\subsection{Multi-scale Structural Anchor Configuration}
\label{app:anchor}

The structural anchor aggregates temporal information across three scales, specified by window lengths in time steps rather than fixed calendar periods.
For ETTh1, ETTh2, Weather, and Electricity, we use window lengths $\{12,24,48\}$ with corresponding strides $\{6,12,24\}$.
For ETTm1 and ETTm2, we use window lengths $\{24,48,96\}$ with corresponding strides $\{12,24,48\}$.
At each scale, each window is encoded by a scale-specific two-layer MLP with a GELU activation between the layers.
The resulting features are averaged across windows within each scale, rescaled by a learnable scalar gain $\alpha_k$, and combined using softmax-normalized weights $\pi_k$ derived from learnable logits $\beta_k$, yielding a single structural anchor token.

\section{Full Long-term Forecasting Results}
\label{app:longterm}

Table~\ref{tab:app-longterm-full} reports horizon-wise multivariate long-term forecasting results corresponding to the averaged scores in the main paper.

\section{Few-shot Forecasting Protocol and Full Results}
\label{app:fewshot}

\subsection{Protocol}
In the few-shot setting, we evaluate performance after training on $L$ initial observations plus the first $10\%$ or $5\%$ of the remaining training data.
For baselines with official few-shot scripts, we follow the released configurations.
Otherwise, we retain the corresponding full-shot configuration, except for the reduced training data and an early-stopping patience of 10 epochs.
CLAIR follows the same training-data restriction.
Under the $5\%$ setting, results for $H{=}720$ on ETTh1 and ETTh2 are omitted due to insufficient training samples for the input--output window, consistent with prior protocols such as OFA~\citep{Zhou2023OneFitsAll}.

\subsection{Full Results}
Tables~\ref{tab:app-fewshot-10-all} and~\ref{tab:app-fewshot-5-full} report horizon-wise few-shot forecasting results corresponding to the averaged scores in the main paper.

\section{Full Short-term Forecasting Results}
\label{app:shortterm}

Table~\ref{tab:short_term_forcasting_results_full} reports short-term forecasting results on the M4 benchmark corresponding to the averaged scores in the main paper.

\section{Sensitivity Analysis}
\label{app:sensitivity}
\begin{figure*}[t]
  \centering
  \includegraphics[width=\textwidth]{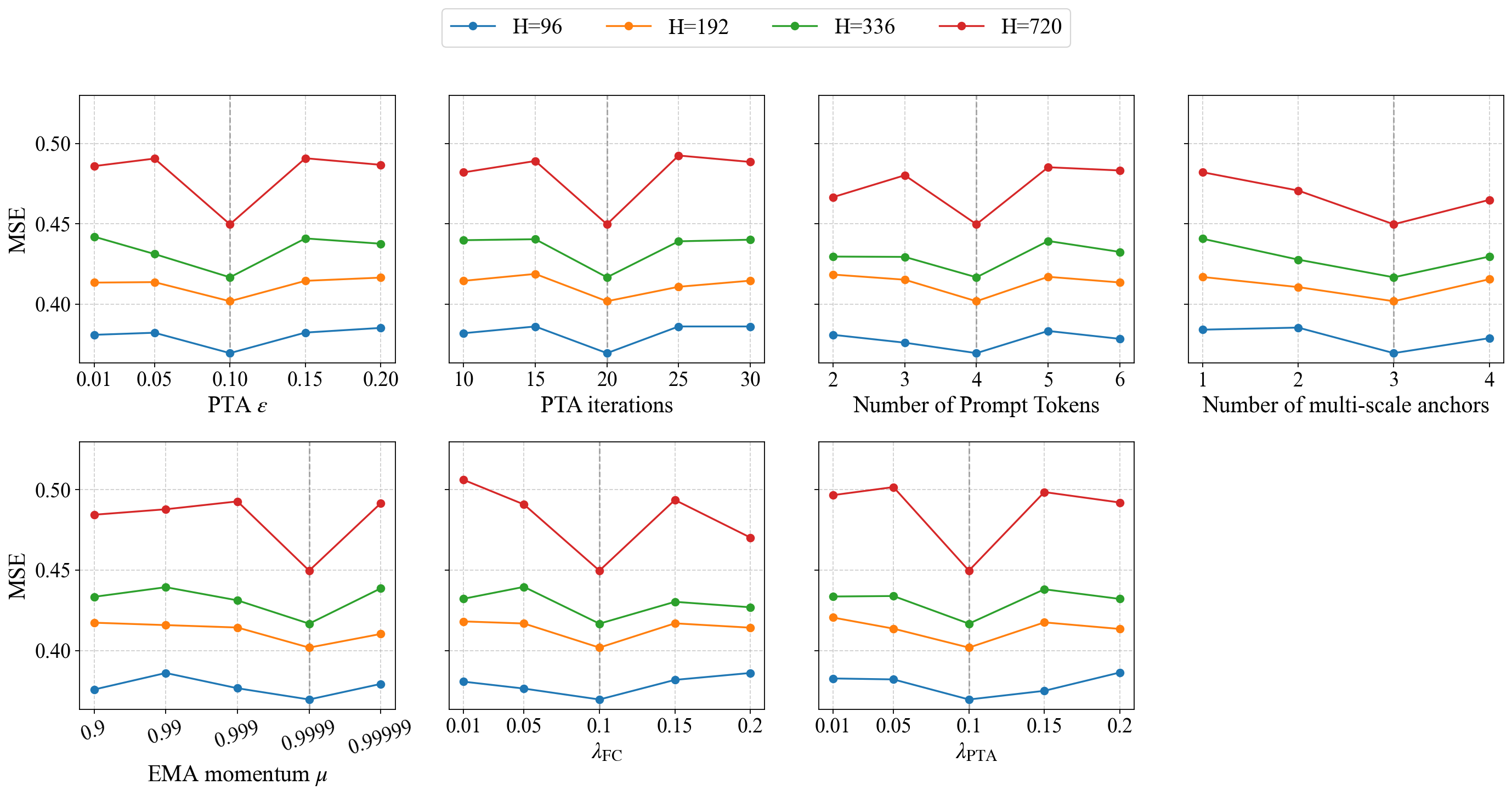}
  \caption{Sensitivity of ETTh1 MSE to seven hyperparameters. Top row, from left to right: PTA
$\epsilon$, PTA iterations, number of prompt tokens, and number of scales in the structural anchor.
Bottom row, from left to right: EMA momentum $\mu$, $\lambda_{\mathrm{FC}}$, and
$\lambda_{\mathrm{PTA}}$. Vertical dashed lines mark the selected values.}
  \label{fig:sensitivity_study}
\end{figure*}

Figure~\ref{fig:sensitivity_study} summarizes CLAIR's hyperparameter sensitivity on ETTh1 with $L{=}512$ and $H\in\{96,192,336,720\}$.
We evaluate model sensitivity to two categories of hyperparameters:
\emph{architectural and training parameters} (the soft-prompt length, the number of structural-anchor scales, the EMA momentum coefficient~$\mu$, and the auxiliary loss weights~$\lambda_{\mathrm{PTA}}$ and~$\lambda_{\mathrm{FC}}$)
and \emph{optimal-transport parameters} (the Sinkhorn entropic regularization~$\varepsilon_{\mathrm{OT}}$ and the number of Sinkhorn iterations).

\subsection{Number of Soft Prompt Tokens}
Varying $n_p$ shows that $n_p{=}4$ consistently provides a strong trade-off.
Fewer prompts limit task-level conditioning capacity, whereas substantially larger values introduce redundant prompt tokens and can degrade accuracy.
We therefore adopt $n_p{=}4$ as the default.

\subsection{Number of Scales in the Structural Anchor}

Starting from the ordered candidate set $\{12,24,48,96\}$, we construct the structural anchor using one to four temporal scales.
Three scales yield the best overall performance: performance degrades with fewer scales, while adding a fourth scale provides little or no additional gain.
Accordingly, we use $K{=}3$ by default, with dataset-specific window sets as described in \ref{app:anchor}.

\subsection{Sinkhorn Regularization and Iterations}
An intermediate $\varepsilon_{\mathrm{OT}}{=}0.1$ balances alignment sharpness and numerical stability.
Twenty Sinkhorn iterations provide the lowest MSE among the tested settings, while increasing the iteration count further does not improve performance.

\subsection{EMA Momentum Coefficient $\mu$}
We evaluate the effect of the EMA momentum coefficient $\mu$ on ETTh1 while keeping all other hyperparameters fixed. $\mu=0.9999$ consistently achieves the lowest MSE across all prediction horizons. Both smaller momentum values ($\mu \leq 0.999$) and the excessively large value $\mu=0.99999$ result in degraded performance. This suggests that a moderate EMA update rate provides a stable yet sufficiently responsive teacher representation. These results support our choice of $\mu=0.9999$ as the default setting.

\subsection{Auxiliary Loss Weights $\lambda_{\mathrm{PTA}}$ and $\lambda_{\mathrm{FC}}$}
Both $\lambda_{\mathrm{FC}}$ and $\lambda_{\mathrm{PTA}}$ show a consistent trend across prediction horizons on ETTh1, with $0.1$ providing the best overall performance for both weights. Smaller or larger values generally lead to higher errors, indicating the importance of appropriately balancing the auxiliary objectives with the primary forecasting loss. We therefore use $\lambda_{\mathrm{FC}}=\lambda_{\mathrm{PTA}}=0.1$ on ETTh1, while the dataset-specific settings for the other benchmarks are provided in \ref{app:hyper}.

\section{Stability and Alignment Analysis}

Beyond forecasting accuracy, we analyze whether CLAIR stabilizes the adaptation process from two complementary perspectives: latent-space stability and output-level robustness. We first examine whether the EMA teacher provides a temporally stable reference and whether PTA keeps student representations aligned with this reference. We then evaluate prediction sensitivity under input perturbations.

\paragraph{Reference Stability and Patch-level Alignment}
We first examine whether the proposed teacher-guided regularization provides a stable reference and maintains student--teacher representation alignment during training. All diagnostics are computed on a fixed validation batch using clean inputs in evaluation mode.

\begin{figure}[t]
\centering
\begin{subfigure}[b]{0.48\columnwidth}
    \centering
    \includegraphics[width=\linewidth]{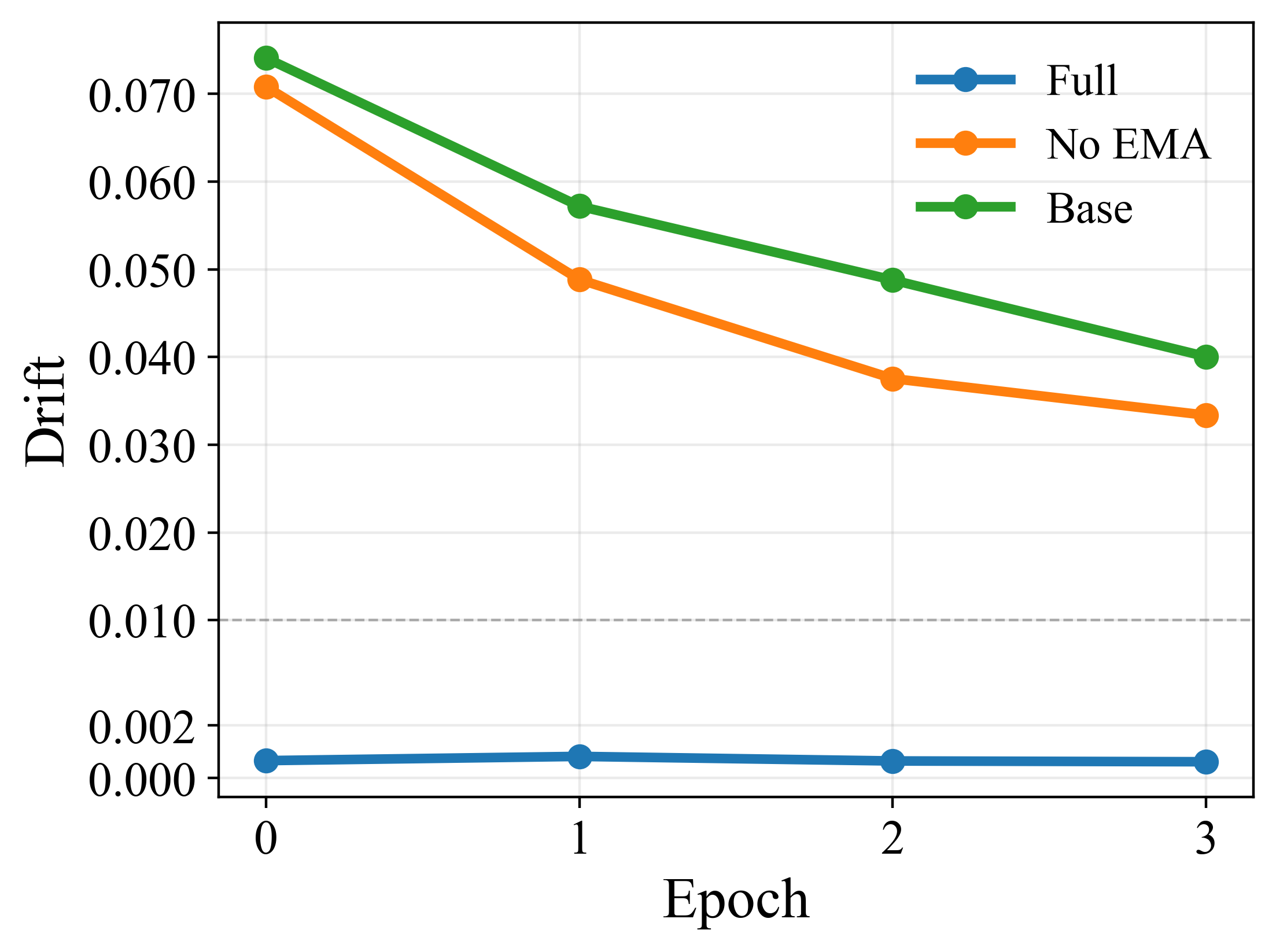}
    \caption{Reference drift}
    \label{fig:reference_drift}
\end{subfigure}%
\hspace{0.02\columnwidth}%
\begin{subfigure}[b]{0.48\columnwidth}
    \centering
    \includegraphics[width=\linewidth]{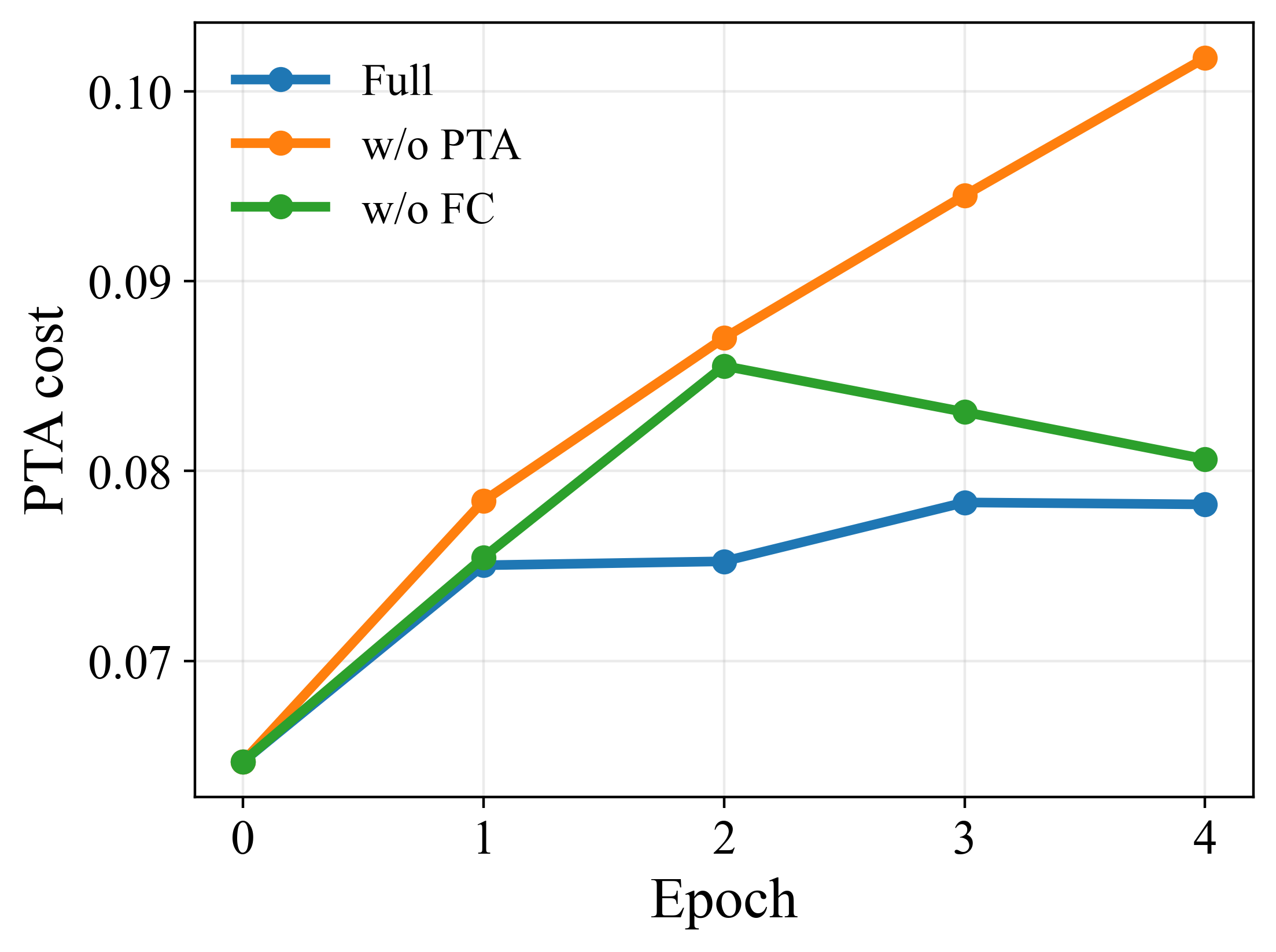}
    \caption{Student--reference PTA cost}
    \label{fig:pta_cost}
\end{subfigure}
\caption{
Reference stability and patch-level alignment during training.
(a) Reference drift on a fixed validation batch.
(b) Student--reference PTA cost.
Lower values indicate a more stable reference and better patch-level alignment, respectively.
}
\label{fig:reference_alignment}
\end{figure}

To measure reference stability, we compute the epoch-to-epoch drift of normalized patch-level reference hidden states:
\begin{equation}
\Delta_{\mathrm{ref}}(e)
=
\frac{1}{UN}
\sum_{u=1}^{U}
\sum_{i=1}^{N}
\left\|
\tilde{h}_{\mathrm{ref},i}^{(u,e)}
-
\tilde{h}_{\mathrm{ref},i}^{(u,e-1)}
\right\|_2 ,
\end{equation}
where $U=BM$ and $N$ denotes the number of patch tokens. For the full model, the reference is given by the EMA teacher. For non-EMA variants, we use the stop-gradient student representation as the reference.
As shown in Figure~\ref{fig:reference_alignment}(a), the full model exhibits substantially lower reference drift than the No EMA and Base variants. This supports the role of the EMA teacher as a temporally stable alignment reference during adaptation.

We further measure whether student patch representations remain aligned with the EMA teacher by computing the student--reference PTA cost:
\begin{equation}
\mathcal{C}_{\mathrm{PTA}}(e)
=
\frac{1}{U}
\sum_{u=1}^{U}
\left\langle
\Pi^{\star(u,e)},
C^{(u,e)}
\right\rangle ,
\end{equation}
where $C^{(u,e)}$ is the student--teacher patch cost matrix and $\Pi^{\star(u,e)}$ is the corresponding Sinkhorn transport plan.
Figure~\ref{fig:reference_alignment}(b) shows that the full model maintains a lower PTA cost than the w/o PTA variant. Without PTA, the student--reference transport cost increases as training proceeds, suggesting that student patch representations drift farther from the EMA teacher reference. The w/o FC variant also shows a higher PTA cost than the full model, suggesting that FC may indirectly support representation-level stability. Together, these results show that the EMA teacher provides a stable reference, while PTA helps maintain patch-level alignment with that reference.

\begin{figure}[t]
\centering
\includegraphics[width=\linewidth]{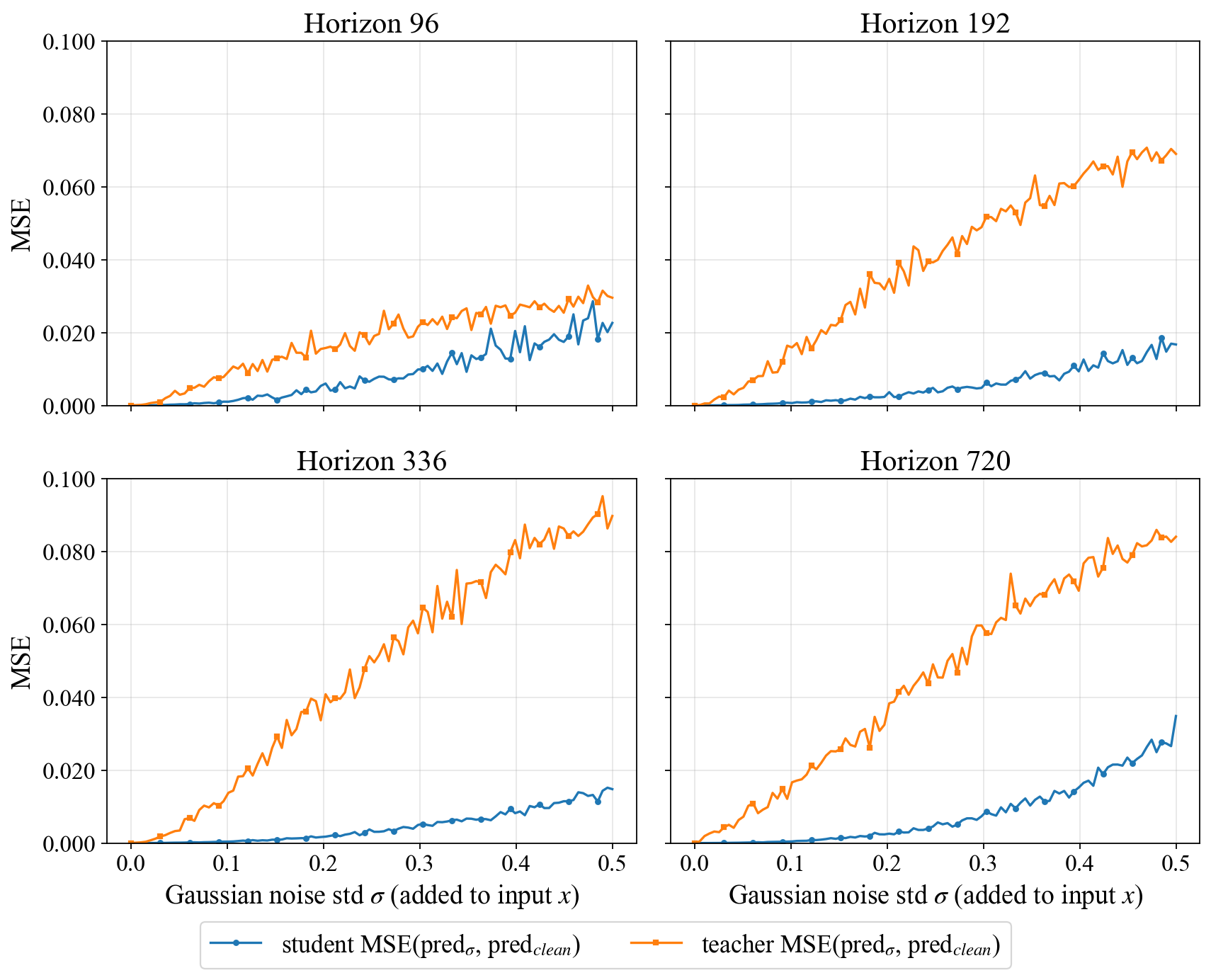}
\caption{
Noise sensitivity of student and EMA teacher predictions. Prediction drift MSE is measured between clean-input and noisy-input predictions across forecasting horizons. Lower values indicate stronger output stability under input perturbations.
}
\label{fig:noise_sensitivity}
\end{figure}

\paragraph{Noise Sensitivity of Predictions} We analyze output-level stability under input perturbations. Given Gaussian noise $\epsilon_{\sigma}\sim\mathcal{N}(0,\sigma^2)$, we construct a noisy input
\begin{equation}
X_{\sigma}
=
X
+
\epsilon_{\sigma},
\end{equation}
and measure prediction drift as
\begin{equation}
D_{\mathrm{drift}}(\sigma)
=
\frac{1}{BHM}
\left\|
\hat{Y}(X_{\sigma})
-
\hat{Y}(X)
\right\|_2^2 .
\end{equation}
This metric measures sensitivity to input noise rather than forecasting error against the ground-truth target.
As shown in Figure~\ref{fig:noise_sensitivity}, the student predictions exhibit lower drift than the EMA teacher predictions across prediction horizons. This suggests that although the EMA teacher serves as a stable clean-view reference during training, the student learns a more perturbation-tolerant forecasting function. The gap becomes larger as the noise level increases, indicating that the asymmetric teacher--student design helps improve output stability under input perturbations.

\section{Additional Diagnostics for Unconstrained PTA}
\label{app:pta_diagnostics}

\subsection{Post-hoc Temporally Constrained OT}
\label{app:pta_constrained_ot}

\begin{figure}[t]
  \centering
  \includegraphics[width=\linewidth]{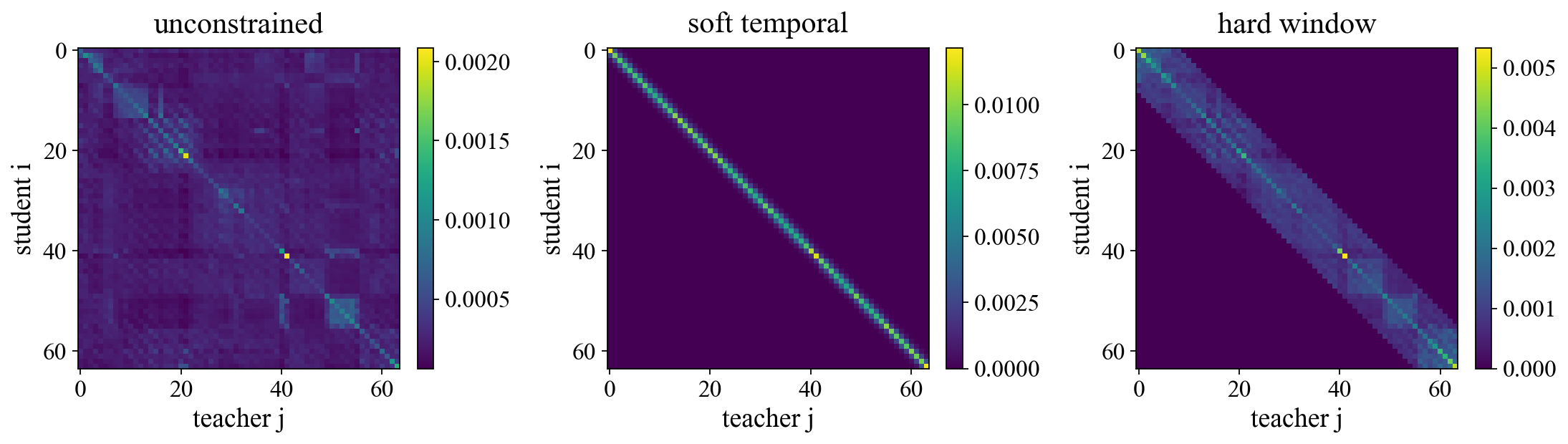}
  \caption{Mean transport plans on ETTh2 at forecast horizon $H=96$ under three alignment settings: unconstrained (left), soft temporal (middle), and hard-window (right). All panels are computed on the first test window used in Table~\ref{tab:pta-etth2-constrained}. The unconstrained panel therefore differs from the multi-window mean plan in Figure~\ref{fig:pta-mean-pi-h96}.}
  \label{fig:pta-etth2-cons-plans}
\end{figure}

\begin{figure}[t]
  \centering
  \includegraphics[width=0.92\linewidth]{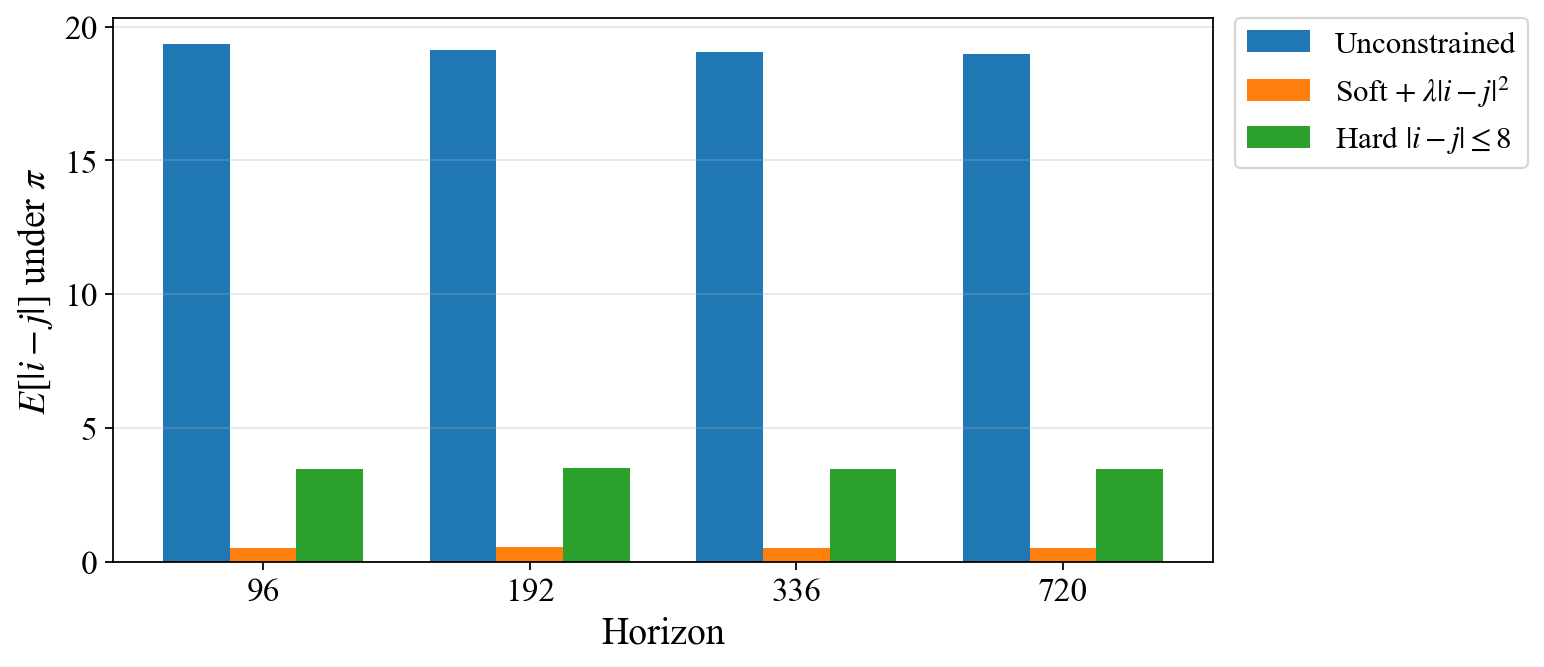}
  \caption{ETTh2: Post-hoc expected transport lag $\mathbb{E}_\pi[|i{-}j|]$ for unconstrained PTA, soft temporal penalty ($C^{\mathrm{soft}}_{ij}=C_{ij}+\lambda|i{-}j|^2$), and hard local window ($|i{-}j|{\le}8$). Bars follow the post-hoc protocol of Table~\ref{tab:pta-etth2-constrained} (first test window). Unconstrained values therefore differ slightly from Table~\ref{tab:pta-etth2-locality}.}
  \label{fig:pta-etth2-cons-lag}
\end{figure}

\begin{table}[t]
\centering
\caption{ETTh2: post-hoc constrained OT locality at H=96 (other horizons are similar). All three rows are recomputed from the cost matrices of the first test window (7 channel sequences) in evaluation mode, using the same protocol as Figures~\ref{fig:pta-etth2-cons-plans}
and~\ref{fig:pta-etth2-cons-lag}.}

\label{tab:pta-etth2-constrained}
\begin{tabular}{lcc}
\toprule
OT variant & $\mathbb{E}[|i{-}j|]$ & Diag.\ mass \\
\midrule
Unconstrained & $19.36$ & $0.048$ \\
Soft $+\,0.05\,|i{-}j|^2$ & $0.52$ & $0.562$ \\
Hard $|i{-}j|{\le}8$ & $3.48$ & $0.147$ \\
\bottomrule
\end{tabular}
\end{table}

We further examine how explicit temporal constraints alter the transport geometry induced by the learned patch representations. Using fixed ETTh2 full-model checkpoints, we recompute the Sinkhorn plans without retraining the forecasting model. We compare unconstrained PTA with a soft temporal penalty,
\begin{equation}
    C_{ij}^{\mathrm{soft}}
    =
    C_{ij}
    +
    \lambda |i-j|^{2},
\end{equation}
and a hard local window that permits transport only when
$|i-j|\leq 8$.
Figure~\ref{fig:pta-etth2-cons-lag} shows that unconstrained PTA retains substantially larger expected patch-index displacement than either temporally constrained alternative across forecasting horizons. The corresponding plans in Figure~\ref{fig:pta-etth2-cons-plans} show that the soft penalty concentrates transport near matching indices,
whereas the hard window produces band-limited correspondence while retaining flexibility within the permitted range.
Table~\ref{tab:pta-etth2-constrained} provides the corresponding locality statistics at $H=96$. Because these post-hoc plans are computed on a single test window, the unconstrained values in Table~\ref{tab:pta-etth2-constrained} ($\mathbb{E}[|i{-}j|]$ = 19.36, diagonal mass 0.048) differ slightly from the multi-window averages in Table~\ref{tab:pta-etth2-locality} (19.19 at H = 96). Together, these post-hoc results place unconstrained PTA between uniform matching and explicitly localized transport. It exhibits a modest temporal preference but retains considerably more non-local correspondence than the constrained alternatives. Because the transport plans are recomputed from fixed representations without retraining, this analysis characterizes correspondence geometry rather than forecasting performance. The soft-penalty setting is therefore treated as a locality stress test rather than as a tuned replacement for unconstrained PTA.

\begin{table}[t]
\centering
\caption{
Trainable parameters (in millions) averaged over prediction
horizons $H \in \{96,192,336,720\}$.
}
\label{tab:param_efficiency}
\small
\setlength{\tabcolsep}{4pt}
\begin{tabular}{lcccc}
\toprule
Method & ETTh1/h2 & ETTm1/m2 & ECL & Weather \\
\midrule
Time-LLM                   & 55.41 & 55.50 & 55.41  & 55.41 \\
S\textsuperscript{2}IP-LLM & 57.04 & 58.50 & 57.04  & 57.04 \\
OFA                        & 17.33 & 9.34  & 17.33  & 17.33 \\
CALF                       & 18.53 & 18.53 & 18.53  & 18.53 \\
FSCA                       & 31.75 & 35.60 & 31.32  & 30.33 \\
SE-LLM                     & 97.33 & 97.33 & 100.31 & 98.61 \\
\midrule
CLAIR                      & 19.18 & 19.24 & 19.18  & 19.18 \\
\bottomrule
\end{tabular}
\end{table}

\section{Additional Diagnostic Analyses}
\label{app:diagnostics}

This section provides extended diagnostic analyses and profiling to substantiate the teacher-guided design. We analyze the parameter and inference efficiency of CLAIR. The EMA teacher receives no gradient updates and is discarded at inference. Thus, trainable and inference-time parameter counts include only the student branch. CLAIR uses substantially fewer trainable parameters than Time-LLM and S\textsuperscript{2}IP-LLM and remains comparable to CALF, although OFA is lighter under this setting. The EMA teacher and Sinkhorn transport introduce training-time overhead, requiring $1.44\times$ training time per iteration and $1.24\times$ peak GPU memory compared with the student-only base. However, these components are removed at inference, where CLAIR uses only the student branch and achieves $1.00\times$ inference latency relative to the student-only base.

\subsection{Profiling Setup}
\label{sec:cost-efficiency}

Teacher-guided training, including the EMA teacher, Sinkhorn-based PTA, and forecast consistency, incurs additional computational cost only during training, with no overhead at inference time. At deployment CLAIR evaluates the student branch alone: input perturbation, teacher forwards, Sinkhorn transport, and the consistency loss are disabled, and teacher weights can be deleted from memory. We profile this claim on the ETTm1 Full checkpoint without retraining ($L{=}512$, $M{=}7$, GPT-2 truncated to 6 layers, patch size $P{=}16$, stride $S{=}8$; multi-scale patches of sizes $24/48/96$ with strides $12/24/48$; and $\lambda_{\mathrm{PTA}}{=}0.10$ and $\lambda_{\mathrm{FC}}{=}0.05$). Unless noted, training steps use batch size $128$ and inference follows the test loader (batch size $1$). Timings are median CUDA-event measurements after warmup on an NVIDIA L40S.

\subsection{Trainable vs.\ stored parameters}
\label{sec:cost-params}

\begin{figure}[t]
  \centering
  \includegraphics[width=\columnwidth]{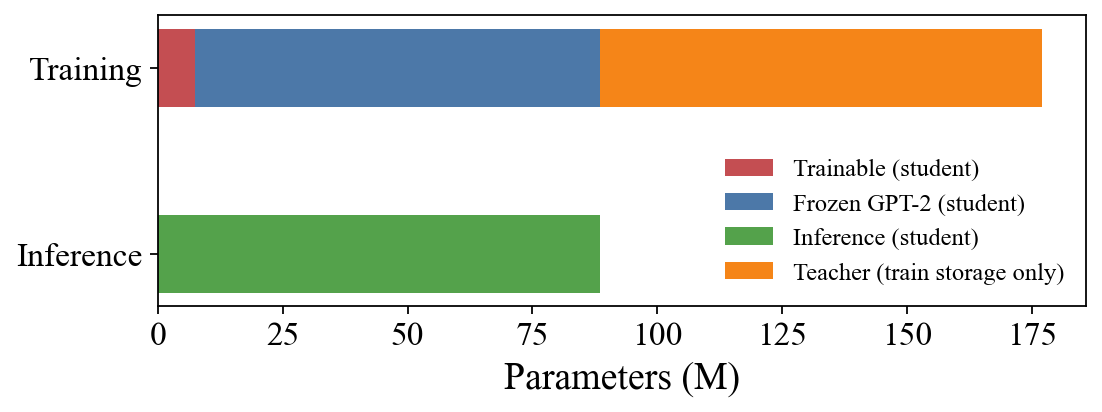}
  \caption{Parameter composition at $H{=}96$ on ETTm1.
  Training retains trainable student parameters (red), frozen GPT-2 weights (blue), and a separate EMA teacher copy (orange).
  At inference, the teacher is discarded and only the student is retained (green).}
  \label{fig:cost-params}
\end{figure}

We distinguish gradient-updated student parameters from the total model weights stored during training and inference.
As shown in Table~\ref{tab:param_efficiency}, CLAIR uses $19.18$--$19.24$M trainable parameters across the evaluated dataset configurations, averaged over $H\in\{96,192,336,720\}$.
For ETTm1, the horizon-averaged count is $19.24$M, whereas the smaller forecasting head at $H{=}96$ reduces the number of trainable parameters to $7.44$M.
Together with $81.11$M frozen GPT-2 weights, the student stores $88.55$M parameters at this horizon.
The EMA teacher has the same parameter count but receives no gradient updates. Its weights corresponding to the trainable student parameters are updated by EMA.
Figure~\ref{fig:cost-params} makes the distinction between training-time storage and deployed model size explicit.
At $H{=}96$, training retains separate student and teacher copies, corresponding to approximately $177.1$M stored parameters in total.
After discarding the teacher, inference requires only the $88.55$M student model.
Thus, the teacher increases weight storage only during training and does not enlarge the deployed model.
In terms of trainable parameters, CLAIR is comparable to CALF and remains smaller than FSCA, Time-LLM, S$^2$IP-LLM, and SE-LLM, although OFA uses fewer trainable parameters.

\subsection{Student-only inference}
\label{sec:cost-infer}

\begin{table}[t]
\centering
\caption{Inference cost on ETTm1. Student-only deploy discards teacher weights. ``Eval w/ teacher'' is the previous eval path that still ran the teacher and Sinkhorn.}
\label{tab:cost-infer}
\begin{tabular}{@{}lccc@{}}
\toprule
Setting  & Latency & Peak GPU & Rel.\ latency \\
\midrule
Student-only  & 4.39\,ms & 0.46\,GB & $1.00\times$ \\
Eval w/ teacher  & 10.32\,ms & 0.76\,GB & $2.35\times$ \\
\bottomrule
\end{tabular}
\end{table}

\begin{figure}[t]
  \centering
  \includegraphics[width=\columnwidth]{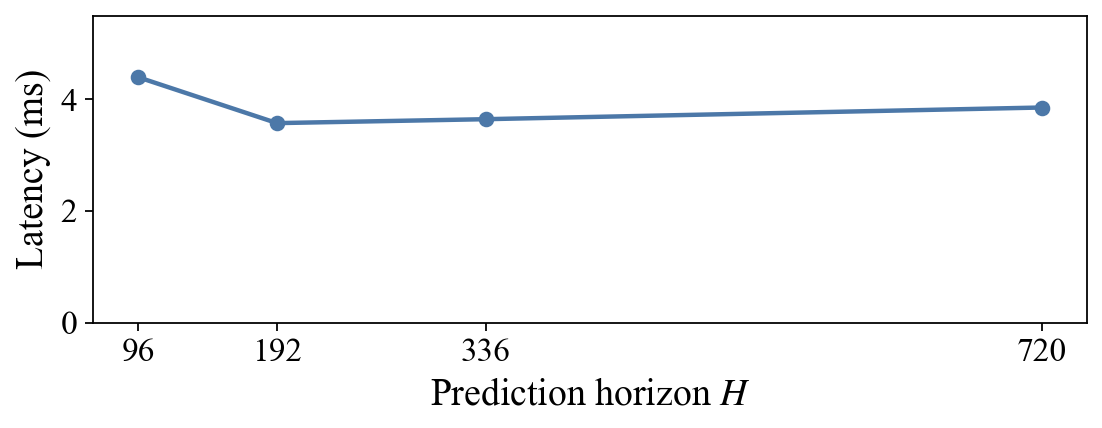}
  \caption{Student-only inference latency versus forecast horizon on ETTm1 (batch $1$). Variation is small relative to the backbone cost.}
  \label{fig:cost-horizon}
\end{figure}

\begin{table}[t]
\centering
\caption{Training cost on ETTm1. Relative columns are vs.\ the student-only Base step.}
\label{tab:cost-train}
\begin{tabular}{@{}lcccc@{}}
\toprule
% Setting & Params & ms/iter & Peak GPU & Rel.\ time / mem \\
Setting & ms/iter & Peak GPU & Rel.\ time / mem \\
\midrule
% CLAIR Full & 19.24M & 1107.7 & 35.73\,GB & $1.44\times$ / $1.24\times$ \\
% Student Base & 19.24M & 771.4 & 28.92\,GB & $1.00\times$ / $1.00\times$ \\
CLAIR Full & 1107.7 & 35.73\,GB & $1.44\times$ / $1.24\times$ \\
Student Base & 771.4 & 28.92\,GB & $1.00\times$ / $1.00\times$ \\
\bottomrule
\end{tabular}
\end{table}

Returned forecasts have always been produced by the student. The profiled change is that evaluation no longer executes the teacher or Sinkhorn.
With teacher weights discarded, ETTm1 inference at $H{=}96$ takes $4.39$\,ms/sample and $0.46$\,GB peak memory (Table~\ref{tab:cost-infer}), matching the student-only Base architecture ($\mathbf{1.00\times}$ latency).
If the teacher forward is left in the eval graph, latency rises to $10.32$\,ms and peak memory to $0.76$\,GB, about $2.35\times$ slower, which illustrates why the skip must be in the forward path rather than only in the reported prediction. Student-forward FLOPs at batch $1$ are $42.35$\,G. Latency stays in a narrow band across $H\in\{96,192,336,720\}$ ($4.39$, $3.57$, $3.64$, and $3.84$\,ms; Figure~\ref{fig:cost-horizon}), because the GPT-2 trunk dominates and only the linear forecast head depends on $H$.

\subsection{Training-time overhead}
\label{sec:cost-train}

\begin{figure}[t]
  \centering
  \includegraphics[width=\columnwidth]{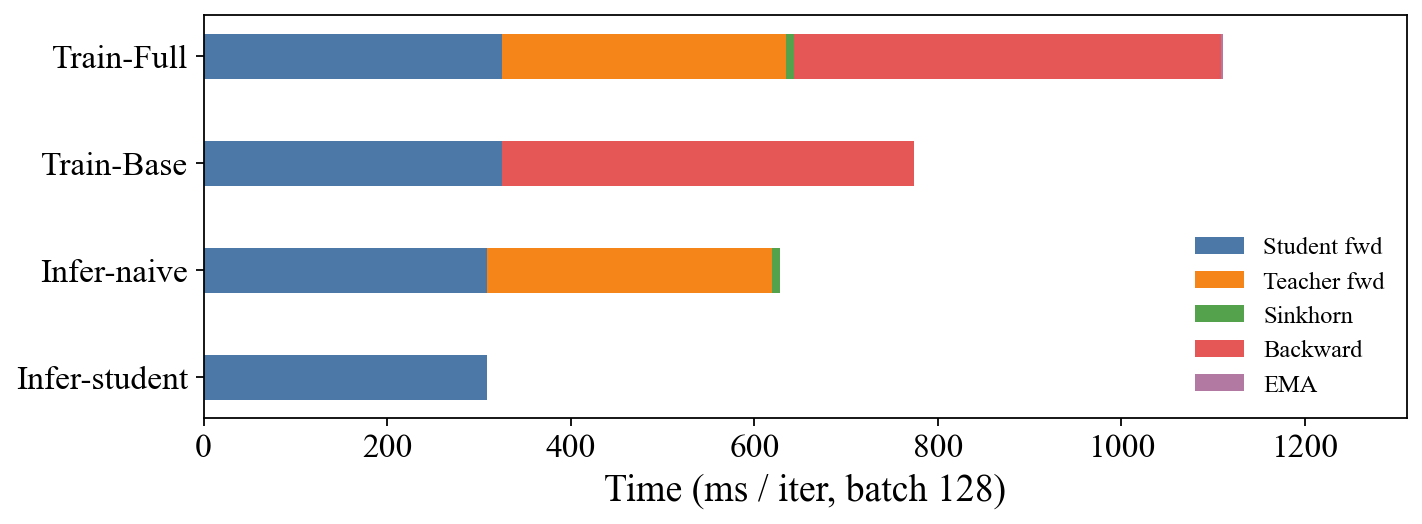}\\[4pt]
  \includegraphics[width=\columnwidth]{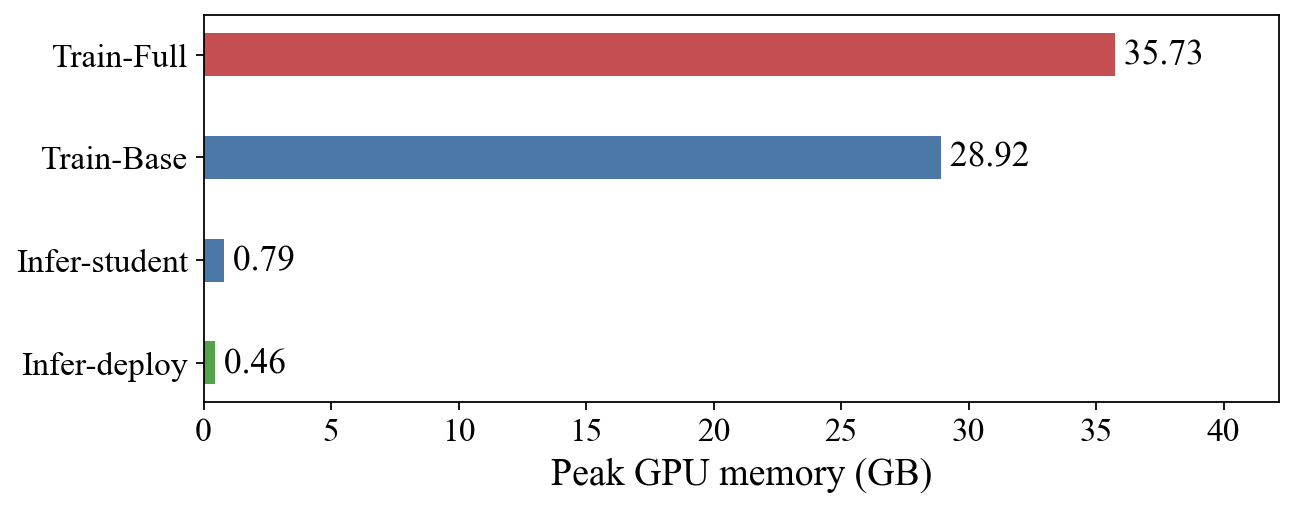}
  \caption{Top: Per-iteration runtime breakdown at batch size $128$ on ETTm1. Train-Full includes the teacher forward pass, Sinkhorn computation, and EMA update, while the teacher forward pass is also retained in the teacher-based evaluation path. Infer-student performs only a single student forward pass. Bottom: Peak GPU memory usage across training and inference configurations. Infer-deploy removes the teacher modules after training.}
  \label{fig:cost-breakdown}
\end{figure}

We compare a Full training step (student forward, no-grad teacher forward, Sinkhorn PTA, FC loss, backward, EMA update) to a student-only Base step (student forward and backward only).
Table~\ref{tab:cost-train} reports $1107.7$\,ms/iter and $35.73$\,GB peak memory for Full versus $771.4$\,ms/iter and $28.92$\,GB for Base, i.e.\ $\mathbf{1.44\times}$ wall-clock and $\mathbf{1.24\times}$ peak GPU memory.
These ratios closely match the ETTh1 and Weather microbenchmarks ($1.44\times$ time; $1.23$--$1.24\times$ memory), indicating that the teacher-guided overhead is structural rather than dataset-specific.
Figure~\ref{fig:cost-breakdown} decomposes the same steps at batch $128$.
The extra Full cost is almost entirely a second (teacher) forward ($\approx 308$\,ms) plus additional backward work on the student graph. Sinkhorn PTA remains cheap: $\approx 8.7$\,ms per iteration ($<1\%$ of a Full step). The additional computation introduced by PTA is modest. Constructing the pairwise student–teacher cost matrices and performing 20 log-domain Sinkhorn iterations for one training step (batch size 128) requires an analytically estimated 8.76 GFLOPs, or about 0.07 GFLOPs per input window. For reference, a single student forward pass requires 42.35 GFLOPs per input window (batch size 1; \ref{sec:cost-infer}).

\begin{figure}[b]
  \centering
  \includegraphics[width=\linewidth]{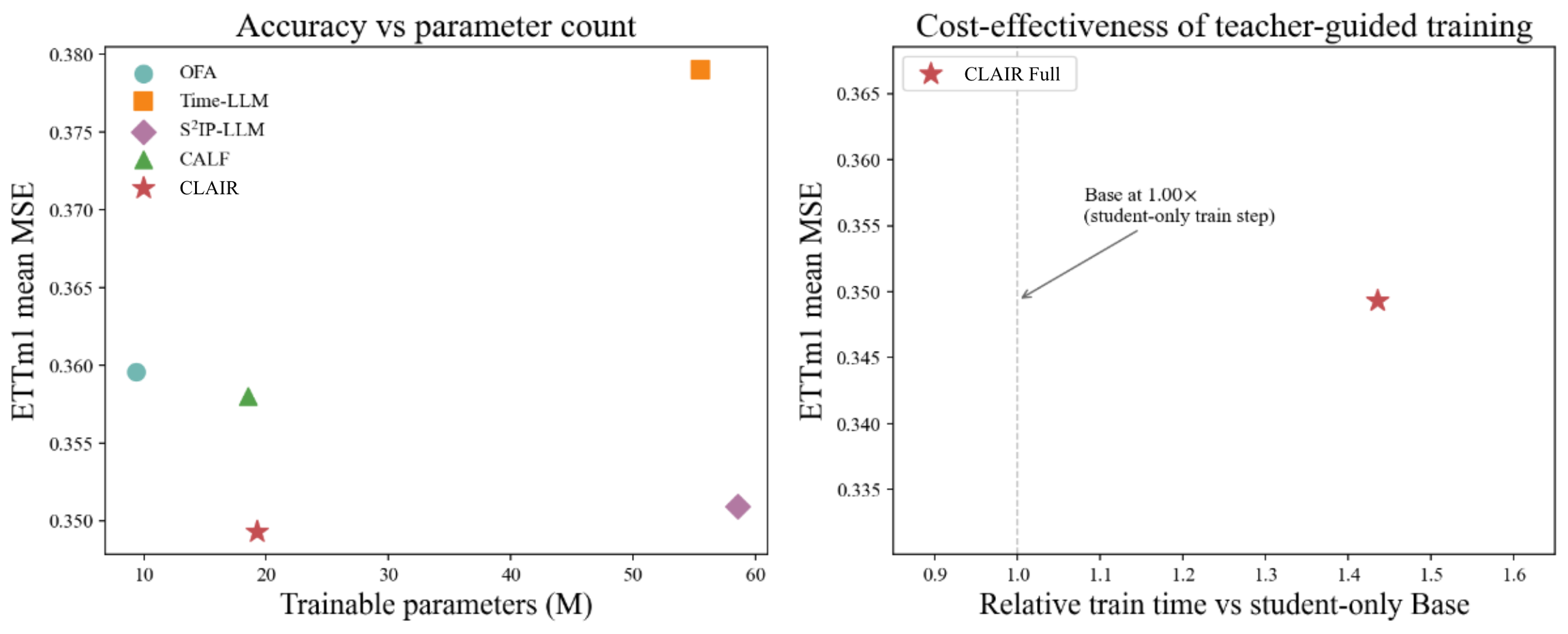}
  \caption{Left: ETTm1 mean MSE versus trainable parameters for LLM-TS baselines and CLAIR. Right: Measured Full versus Base training-time ratio; inference uses the student only in both cases.}
  \label{fig:cost-pareto}
\end{figure}

\subsection{Cost-effectiveness versus LLM baselines}
\label{sec:cost-pareto}
Figure~\ref{fig:cost-pareto} (left) plots ETTm1 mean MSE (full-shot average over four horizons) against trainable parameter count.
CLAIR attains $0.3493$ MSE at $19.24$M trainable parameters, improving on Time-LLM ($0.3790$ / $55.50$M), S$^2$IP-LLM ($0.3509$ / $58.50$M), OFA ($0.3596$ / $9.34$M), and CALF ($0.3580$ / $18.53$M). CLAIR achieves lower horizon-averaged MSE than these four baselines, with a trainable parameter count close to CALF, below Time-LLM and S$^2$IP-LLM, and above OFA. These counts describe trainable parameters. Total stored weights and measured training costs are reported separately in Sections~\ref{sec:cost-params} and \ref{sec:cost-train}. The right panel of Figure~\ref{fig:cost-pareto} places CLAIR's Full training at $1.44\times$ the step time of the student-only Base. This additional cost is incurred only during optimization, since both configurations use the student branch alone at inference.

\section{Run-to-run Variability Analysis}
\label{app:run_variability}

\begin{table}[!t]
\centering
\caption{
Run-to-run variability and approximate 95\% confidence
intervals for selected long-term forecasting comparisons.
}
\label{tab:baseline_welch_ci}

\footnotesize
\setlength{\tabcolsep}{2.5pt}
\renewcommand{\arraystretch}{1.05}

\resizebox{\columnwidth}{!}{%
\begin{tabular}{@{}llcccc@{}}
\toprule
Dataset
& Baseline
& $H$
& \makecell{CLAIR MSE\\$\pm$ SD}
& \makecell{Baseline MSE\\$\pm$ SD}
& \makecell{$\Delta$MSE\\{[95\% CI]}} \\
\midrule

\multirow[t]{4}{*}{Weather}
& \multirow[t]{4}{*}{FSCA}
& 96
& \makecell{$0.1498$\\$\pm 0.00121$}
& \makecell{$0.1492$\\$\pm 0.00090$}
& \makecell{$-0.0006$\\$[-0.0031,\,0.0019]$} \\

& & 192
& \makecell{$0.1923$\\$\pm 0.00044$}
& \makecell{$0.1935$\\$\pm 0.00089$}
& \makecell{$0.0012$\\$[-0.0007,\,0.0031]$} \\

& & 336
& \makecell{$0.2469$\\$\pm 0.00050$}
& \makecell{$0.2474$\\$\pm 0.00188$}
& \makecell{$0.0005$\\$[-0.0038,\,0.0048]$} \\

& & 720
& \makecell{$0.3152$\\$\pm 0.00058$}
& \makecell{$0.3221$\\$\pm 0.00081$}
& \makecell{$0.0069$\\$[0.0052,\,0.0086]$} \\

\midrule

\multirow[t]{4}{*}{ETTh2}
& \multirow[t]{4}{*}{FSCA}
& 96
& \makecell{$0.2840$\\$\pm 0.00055$}
& \makecell{$0.2886$\\$\pm 0.00168$}
& \makecell{$0.0046$\\$[0.0009,\,0.0083]$} \\

& & 192
& \makecell{$0.3530$\\$\pm 0.00060$}
& \makecell{$0.3571$\\$\pm 0.00010$}
& \makecell{$0.0041$\\$[0.0027,\,0.0055]$} \\

& & 336
& \makecell{$0.3715$\\$\pm 0.00039$}
& \makecell{$0.3797$\\$\pm 0.00087$}
& \makecell{$0.0082$\\$[0.0064,\,0.0100]$} \\

& & 720
& \makecell{$0.4171$\\$\pm 0.00051$}
& \makecell{$0.4134$\\$\pm 0.00289$}
& \makecell{$-0.0037$\\$[-0.0106,\,0.0032]$} \\

\midrule

\multirow[t]{4}{*}{ETTm1}
& \multirow[t]{4}{*}{S$^2$IP-LLM}
& 96
& \makecell{$0.2902$\\$\pm 0.00021$}
& \makecell{$0.2928$\\$\pm 0.00213$}
& \makecell{$0.0026$\\$[-0.0026,\,0.0078]$} \\

& & 192
& \makecell{$0.3330$\\$\pm 0.00202$}
& \makecell{$0.3324$\\$\pm 0.00237$}
& \makecell{$-0.0006$\\$[-0.0056,\,0.0044]$} \\

& & 336
& \makecell{$0.3616$\\$\pm 0.00254$}
& \makecell{$0.3625$\\$\pm 0.00200$}
& \makecell{$0.0009$\\$[-0.0044,\,0.0062]$} \\

& & 720
& \makecell{$0.4125$\\$\pm 0.00262$}
& \makecell{$0.4160$\\$\pm 0.00305$}
& \makecell{$0.0035$\\$[-0.0030,\,0.0100]$} \\

\midrule

\multirow[t]{4}{*}{ETTm2}
& \multirow[t]{4}{*}{FSCA}
& 96
& \makecell{$0.1671$\\$\pm 0.00027$}
& \makecell{$0.1675$\\$\pm 0.00144$}
& \makecell{$0.0004$\\$[-0.0030,\,0.0038]$} \\

& & 192
& \makecell{$0.2230$\\$\pm 0.00050$}
& \makecell{$0.2301$\\$\pm 0.00279$}
& \makecell{$0.0071$\\$[0.0005,\,0.0137]$} \\

& & 336
& \makecell{$0.2740$\\$\pm 0.00059$}
& \makecell{$0.2752$\\$\pm 0.00277$}
& \makecell{$0.0012$\\$[-0.0053,\,0.0077]$} \\

& & 720
& \makecell{$0.3609$\\$\pm 0.00037$}
& \makecell{$0.3594$\\$\pm 0.00355$}
& \makecell{$-0.0015$\\$[-0.0102,\,0.0072]$} \\

\bottomrule
\end{tabular}%
}
\end{table}

Among the six long-term forecasting benchmarks, CLAIR shows
relatively small improvements in horizon-averaged MSE over
the strongest evaluated LLM baseline on Weather, ETTh2,
ETTm1, and ETTm2.
To assess these modest performance differences against
run-to-run variability, we conduct an additional analysis
on these four datasets.
For each dataset, we select the LLM baseline with the lowest
horizon-averaged MSE and report the mean and standard
deviation (SD) over three runs at each forecasting horizon.
Table~\ref{tab:baseline_welch_ci} also presents approximate
95\% confidence intervals (CIs) for the MSE differences,
computed using Welch's method.
The difference is defined as
$\Delta\mathrm{MSE}
=\mathrm{MSE}_{\mathrm{baseline}}-\mathrm{MSE}_{\mathrm{CLAIR}}$,
so positive values favor CLAIR.
CLAIR achieves lower mean MSE in 12 of the 16 comparisons,
with five confidence intervals entirely above zero.
The remaining intervals include zero, indicating that
the evidence for improvements varies across datasets
and forecasting horizons.
These results characterize the uncertainty associated
with the modest performance differences observed in
the selected long-term forecasting comparisons.

\bibliographystyle{elsarticle-num}
% \twocolumn
\bibliography{refs}

@inproceedings{Grill2020BYOL,
  author    = {Jean{-}Bastien Grill and
               Florian Strub and
               Florent Altch{\'{e}} and
               Corentin Tallec and
               Pierre H. Richemond and
               Elena Buchatskaya and
               Carl Doersch and
               Bernardo {\'{A}}vila Pires and
               Zhaohan Guo and
               Mohammad Gheshlaghi Azar and
               Bilal Piot and
               Koray Kavukcuoglu and
               R{\'{e}}mi Munos and
               Michal Valko},
  title     = {Bootstrap Your Own Latent - {A} New Approach to Self-Supervised Learning},
  booktitle = {Advances in Neural Information Processing Systems 33: Annual Conference
               on Neural Information Processing Systems 2020, NeurIPS 2020, December
               6-12, 2020, virtual},
  year      = {2020},
}

@inproceedings{Chen2021SimSiam,
  author    = {Xinlei Chen and
               Kaiming He},
  title     = {Exploring Simple Siamese Representation Learning},
  booktitle = {{IEEE} Conference on Computer Vision and Pattern Recognition, {CVPR}
               2021, virtual, June 19-25, 2021},
  pages     = {15750--15758},
  year      = {2021},
}

@inproceedings{Zhou2023OneFitsAll,
  author    = {Tian Zhou and
               Peisong Niu and
               Xue Wang and
               Liang Sun and
               Rong Jin},
  title     = {One Fits All: Power General Time Series Analysis by Pretrained LM},
  booktitle = {Advances in Neural Information Processing Systems 36: Annual Conference
               on Neural Information Processing Systems 2023, NeurIPS 2023, New Orleans,
               LA, USA, December 10 - 16, 2023},
  year      = {2023},
}

@article{Chang2025LLM4TS,
  author       = {Ching Chang and
                  Wei{-}Yao Wang and
                  Wen{-}Chih Peng and
                  Tien{-}Fu Chen},
  title        = {{LLM4TS:} Aligning Pre-Trained LLMs as Data-Efficient Time-Series
                  Forecasters},
  journal      = {{ACM} Trans. Intell. Syst. Technol.},
  volume       = {16},
  number       = {3},
  pages        = {60:1--60:20},
  year         = {2025},
}

@inproceedings{Jin2024TIMELLM,
  author    = {Ming Jin and
               Shiyu Wang and
               Lintao Ma and
               Zhixuan Chu and
               James Y. Zhang and
               Xiaoming Shi and
               Pin{-}Yu Chen and
               Yuxuan Liang and
               Yuan{-}Fang Li and
               Shirui Pan and
               Qingsong Wen},
  title     = {Time-LLM: Time Series Forecasting by Reprogramming Large Language
               Models},
  booktitle = {The Twelfth International Conference on Learning Representations,
               {ICLR} 2024, Vienna, Austria, May 7-11, 2024},
  year      = {2024},
}

@inproceedings{Pan2024S2IP,
  author       = {Zijie Pan and
                  Yushan Jiang and
                  Sahil Garg and
                  Anderson Schneider and
                  Yuriy Nevmyvaka and
                  Dongjin Song},
  title        = {{S2IP-LLM:} Semantic Space Informed Prompt Learning with {LLM} for
                  Time Series Forecasting},
  booktitle    = {Forty-first International Conference on Machine Learning, {ICML} 2024,
                  Vienna, Austria, July 21-27, 2024},
  series       = {Proceedings of Machine Learning Research},
  volume       = {235},
  pages        = {39135--39153},
  year         = {2024},
}

@inproceedings{Liu2025TimeCMA,
  author    = {Chenxi Liu and
               Qianxiong Xu and
               Hao Miao and
               Sun Yang and
               Lingzheng Zhang and
               Cheng Long and
               Ziyue Li and
               Rui Zhao},
  title     = {TimeCMA: Towards LLM-Empowered Multivariate Time Series Forecasting
               via Cross-Modality Alignment},
  booktitle = {Thirty-Ninth {AAAI} Conference on Artificial Intelligence, Thirty-Seventh
               Conference on Innovative Applications of Artificial Intelligence,
               Fifteenth Symposium on Educational Advances in Artificial Intelligence,
               {AAAI} 2025, Philadelphia, PA, USA, February 25 - March 4, 2025},
  pages     = {18780--18788},
  year      = {2025},
}

@inproceedings{Liu2025CALF,
  author    = {Peiyuan Liu and
               Hang Guo and
               Tao Dai and
               Naiqi Li and
               Jigang Bao and
               Xudong Ren and
               Yong Jiang and
               Shu{-}Tao Xia},
  title     = {{CALF:} Aligning LLMs for Time Series Forecasting via Cross-modal
               Fine-Tuning},
  booktitle = {Thirty-Ninth {AAAI} Conference on Artificial Intelligence, Thirty-Seventh
               Conference on Innovative Applications of Artificial Intelligence,
               Fifteenth Symposium on Educational Advances in Artificial Intelligence,
               {AAAI} 2025, Philadelphia, PA, USA, February 25 - March 4, 2025},
  pages     = {18915--18923},
  year      = {2025},
}

@inproceedings{Zhao2024himtm,
  author       = {Shubao Zhao and
                  Ming Jin and
                  Zhaoxiang Hou and
                  Chengyi Yang and
                  Zengxiang Li and
                  Qingsong Wen and
                  Yi Wang},
  editor       = {Edoardo Serra and
                  Francesca Spezzano},
  title        = {HiMTM: Hierarchical Multi-Scale Masked Time Series Modeling with Self-Distillation
                  for Long-Term Forecasting},
  booktitle    = {Proceedings of the 33rd {ACM} International Conference on Information
                  and Knowledge Management, {CIKM} 2024, Boise, ID, USA, October 21-25,
                  2024},
  pages        = {3352--3362},
  year         = {2024},
}

@inproceedings{Zeng2023dlinear,
  author       = {Ailing Zeng and
                  Muxi Chen and
                  Lei Zhang and
                  Qiang Xu},
  title        = {Are Transformers Effective for Time Series Forecasting?},
  booktitle    = {Thirty-Seventh {AAAI} Conference on Artificial Intelligence, {AAAI}
                  2023, Thirty-Fifth Conference on Innovative Applications of Artificial
                  Intelligence, {IAAI} 2023, Thirteenth Symposium on Educational Advances
                  in Artificial Intelligence, {EAAI} 2023, Washington, DC, USA, February
                  7-14, 2023},
  pages        = {11121--11128},
  year         = {2023},
}

@inproceedings{Liu2024itransformer,
  author    = {Yong Liu and
               Tengge Hu and
               Haoran Zhang and
               Haixu Wu and
               Shiyu Wang and
               Lintao Ma and
               Mingsheng Long},
  title     = {iTransformer: Inverted Transformers Are Effective for Time Series
               Forecasting},
  booktitle = {The Twelfth International Conference on Learning Representations,
               {ICLR} 2024, Vienna, Austria, May 7-11, 2024},
  year      = {2024},
}

@inproceedings{Nie2023patchtst,
  author    = {Yuqi Nie and
               Nam H. Nguyen and
               Phanwadee Sinthong and
               Jayant Kalagnanam},
  title     = {A Time Series is Worth 64 Words: Long-term Forecasting with Transformers},
  booktitle = {The Eleventh International Conference on Learning Representations,
               {ICLR} 2023, Kigali, Rwanda, May 1-5, 2023},
  year      = {2023},
}

@inproceedings{Wu2021autoformer,
  author       = {Haixu Wu and
                  Jiehui Xu and
                  Jianmin Wang and
                  Mingsheng Long},
  title        = {Autoformer: Decomposition Transformers with Auto-Correlation for Long-Term
                  Series Forecasting},
  booktitle    = {Advances in Neural Information Processing Systems 34: Annual Conference
                  on Neural Information Processing Systems 2021, NeurIPS 2021, December
                  6-14, 2021, virtual},
  pages        = {22419--22430},
  year         = {2021},
}

@inproceedings{Zhou2021informer,
  author       = {Haoyi Zhou and
                  Shanghang Zhang and
                  Jieqi Peng and
                  Shuai Zhang and
                  Jianxin Li and
                  Hui Xiong and
                  Wancai Zhang},
  title        = {Informer: Beyond Efficient Transformer for Long Sequence Time-Series
                  Forecasting},
  booktitle    = {Thirty-Fifth {AAAI} Conference on Artificial Intelligence, {AAAI}
                  2021, Thirty-Third Conference on Innovative Applications of Artificial
                  Intelligence, {IAAI} 2021, The Eleventh Symposium on Educational Advances
                  in Artificial Intelligence, {EAAI} 2021, Virtual Event, February 2-9,
                  2021},
  pages        = {11106--11115},
  year         = {2021},
}

@article{Makridakis2018m4,
title = {The M4 Competition: Results, findings, conclusion and way forward},
journal = {International Journal of Forecasting},
volume = {34},
number = {4},
pages = {802-808},
year = {2018},
issn = {0169-2070},
author = {Spyros Makridakis and Evangelos Spiliotis and Vassilios Assimakopoulos},
}

@inproceedings{Tarvainen2017MeanTeacher,
  author       = {Antti Tarvainen and
                  Harri Valpola},
  title        = {Mean teachers are better role models: Weight-averaged consistency
                  targets improve semi-supervised deep learning results},
  booktitle    = {Advances in Neural Information Processing Systems 30: Annual Conference
                  on Neural Information Processing Systems 2017, December 4-9, 2017,
                  Long Beach, CA, {USA}},
  pages        = {1195--1204},
  year         = {2017},
}

@article{Skaf2025tsrc,
  author       = {Wadie Skaf and
                  Mitra Baratchi and
                  Holger H. Hoos},
  title        = {Time series representations classroom {(TSRC):} a teacher-student-based
                  framework for interpretability-enhanced unsupervised time series representation
                  learning},
  journal      = {Mach. Learn.},
  volume       = {114},
  number       = {12},
  pages        = {276},
  year         = {2025},
}

@article{Fan2025stas,
  author       = {Jinxiao Fan and
                  Pengfei Wang and
                  Liang Liu and
                  Huadong Ma},
  title        = {Self-Optimizing Teacher and Auto-Matching Student Framework for Change-Point
                  Representation Learning in Time Series Forecasting},
  journal      = {{ACM} Trans. Intell. Syst. Technol.},
  volume       = {16},
  number       = {3},
  pages        = {72:1--72:25},
  year         = {2025},
}

@inproceedings{wu2023timesnet,
  author       = {Haixu Wu and
                  Tengge Hu and
                  Yong Liu and
                  Hang Zhou and
                  Jianmin Wang and
                  Mingsheng Long},
  title        = {TimesNet: Temporal 2D-Variation Modeling for General Time Series Analysis},
  booktitle    = {The Eleventh International Conference on Learning Representations,
                  {ICLR} 2023, Kigali, Rwanda, May 1-5, 2023},
  year         = {2023},
}

@article{radford2019language,
  title={Language Models are Unsupervised Multitask Learners},
  author={Radford, Alec and Wu, Jeff and Child, Rewon and Luan, David and Amodei, Dario and Sutskever, Ilya},
  year={2019}
}

@inproceedings{oreshkin2019nbeats,
  author       = {Boris N. Oreshkin and
                  Dmitri Carpov and
                  Nicolas Chapados and
                  Yoshua Bengio},
  title        = {{N-BEATS:} Neural basis expansion analysis for interpretable time
                  series forecasting},
  booktitle    = {8th International Conference on Learning Representations, {ICLR} 2020,
                  Addis Ababa, Ethiopia, April 26-30, 2020},
  year         = {2020},
}

@inproceedings{Hu2025FSCA,
  author       = {Yuxiao Hu and
                  Qian Li and
                  Dongxiao Zhang and
                  Jinyue Yan and
                  Yuntian Chen},
  title        = {Context-Alignment: Activating and Enhancing LLMs Capabilities in Time
                  Series},
  booktitle    = {The Thirteenth International Conference on Learning Representations,
                  {ICLR} 2025, Singapore, April 24-28, 2025},
  year         = {2025},
}

@inproceedings{Liu2026SELLM,
 author = {Liu, Hao and Zhang, Xiaoxing and Yang, Chun and Zhu, Xiaobin},
 booktitle = {International Conference on Learning Representations},
 editor = {C. Vondrick and B. Hariharan and C. Raffel and L. Pinto and D. Yang and A. Faust},
 pages = {101404--101425},
 title = {Semantic-Enhanced Time-Series Forecasting via Large Language Models},
 volume = {2026},
 year = {2026}
}

@article{Wang2025TSHTFA,
  author       = {Pengfei Wang and
                  Huanran Zheng and
                  Qi'ao Xu and
                  Silong Dai and
                  Yiqiao Wang and
                  Wenjing Yue and
                  Wei Zhu and
                  Tianwen Qian and
                  Liang Zhao},
  title        = {{TS-HTFA:} Advancing Time-Series Forecasting via Hierarchical Text-Free
                  Alignment with Large Language Models},
  journal      = {Symmetry},
  volume       = {17},
  number       = {3},
  pages        = {401},
  year         = {2025},
}

@inproceedings{Vaswani2017Attention,
  author       = {Ashish Vaswani and
                  Noam Shazeer and
                  Niki Parmar and
                  Jakob Uszkoreit and
                  Llion Jones and
                  Aidan N. Gomez and
                  Lukasz Kaiser and
                  Illia Polosukhin},
  title        = {Attention is All you Need},
  booktitle    = {Advances in Neural Information Processing Systems 30: Annual Conference
                  on Neural Information Processing Systems 2017, December 4-9, 2017,
                  Long Beach, CA, {USA}},
  pages        = {5998--6008},
  year         = {2017},
}

@inproceedings{Jiao2020TinyBERT,
  author       = {Xiaoqi Jiao and
                  Yichun Yin and
                  Lifeng Shang and
                  Xin Jiang and
                  Xiao Chen and
                  Linlin Li and
                  Fang Wang and
                  Qun Liu},
  title        = {TinyBERT: Distilling {BERT} for Natural Language Understanding},
  booktitle    = {Findings of the Association for Computational Linguistics: {EMNLP}
                  2020, Online Event, 16-20 November 2020},
  series       = {Findings of {ACL}},
  volume       = {{EMNLP} 2020},
  pages        = {4163--4174},
  year         = {2020},
}

@inproceedings{Cui2025MultiLevelOT,
  author       = {Xiao Cui and
                  Mo Zhu and
                  Yulei Qin and
                  Liang Xie and
                  Wengang Zhou and
                  Houqiang Li},
  title        = {Multi-Level Optimal Transport for Universal Cross-Tokenizer Knowledge
                  Distillation on Language Models},
  booktitle    = {Thirty-Ninth {AAAI} Conference on Artificial Intelligence, Thirty-Seventh
                  Conference on Innovative Applications of Artificial Intelligence,
                  Fifteenth Symposium on Educational Advances in Artificial Intelligence,
                  {AAAI} 2025, Philadelphia, PA, USA, February 25 - March 4, 2025},
  pages        = {23724--23732},
  year         = {2025},
}

@inproceedings{Cuturi2013Sinkhorn,
  author       = {Marco Cuturi},
  title        = {Sinkhorn Distances: Lightspeed Computation of Optimal Transport},
  booktitle    = {Advances in Neural Information Processing Systems 26: 27th Annual
                  Conference on Neural Information Processing Systems 2013. Proceedings
                  of a meeting held December 5-8, 2013, Lake Tahoe, Nevada, United States},
  pages        = {2292--2300},
  year         = {2013},
}

\end{document}